\documentclass[journal]{IEEEtran}
\usepackage{amsmath,amsfonts}
\usepackage{subcaption}					
\usepackage{array}
\usepackage{textcomp}
\usepackage{stfloats}
\usepackage{url}
\usepackage{verbatim}
\usepackage{graphicx}
\usepackage[utf8]{inputenc}
\usepackage[T1]{fontenc}				
\usepackage[english]{babel}				
\usepackage{microtype}					
\usepackage{blindtext}					
\usepackage{booktabs}					
\usepackage{color}						
\usepackage{nomencl}					
\usepackage{float}
\usepackage{caption}					
\usepackage{bm}	
\usepackage{colortbl}
\usepackage{amssymb}
\usepackage{here} 
\definecolor{dunkelgrau}{rgb}{0.8,0.8,0.8}
\definecolor{hellgrau}{rgb}{0.95,0.95,0.95}
\usepackage{epstopdf}
\usepackage{multirow}
\usepackage{algorithm}
\usepackage{algpseudocode} 
\usepackage{siunitx}
\usepackage[numbers,sort&compress]{natbib}

\usepackage{xcolor}
\usepackage{lmodern}
\usepackage{listings}
\usepackage{hyperref}

\def\BibTeX{{\rm B\kern-.05em{\sc i\kern-.025em b}\kern-.08em
    T\kern-.1667em\lower.7ex\hbox{E}\kern-.125emX}}
\usepackage{balance}

\usepackage{amsthm}
\theoremstyle{definition}
\newtheorem{definition}{Definition}[section]

\begin{document}

\title{Multi-Objective Loss Balancing for Physics-Informed Deep Learning
\thanks{The authors would like to thankfully acknowledge the facilities of Design++ at ETH Zürich and the funding through ETH Foundation grant No. 2020-HS-388 (provided by Kollbrunner/Rodio) as well as the SDSC Project "Domain-Aware AI-augmented Design of Bridges (DAAAD Bridges)".}}

\author{
    \IEEEauthorblockN{Rafael Bischof}\\
    \IEEEauthorblockA{\textit{Swiss Data Science Center} \\
    Zürich, Switzerland\\
    rafael.bischof@sdsc.ethz.ch}\\
    
    \vspace{3mm}
    \IEEEauthorblockN{Michael A. Kraus}\\
    \IEEEauthorblockA{\textit{Chair of Concrete Structures and Bridge Design (IBK)/ Design++, ETH Zürich} \\
    Zürich, Switzerland \\
    kraus@ibk.baug.ethz.ch}\\

}

\markboth{Preprint - Version 16.11.2022. Submitted for Review}%
{Transactions on Neural Networks and Learning Systems}

\maketitle

\begin{abstract}
Physics-Informed Neural Networks (PINN) are algorithms from deep learning leveraging physical laws by including partial differential equations together with a respective set of boundary and initial conditions as penalty terms into their loss function.
In this work, we observe the significant role of correctly weighting the combination of multiple competitive loss functions for training PINNs effectively. To this end, we implement and evaluate different methods aiming at balancing the contributions of multiple terms of the PINNs loss function and their gradients. After reviewing of three existing loss scaling approaches (Learning Rate Annealing, GradNorm and SoftAdapt), we propose a novel self-adaptive loss balancing scheme for PINNs named \emph{ReLoBRaLo} (Relative Loss Balancing with Random Lookback). We extensively evaluate the performance of the aforementioned balancing schemes by solving both forward as well as inverse problems on three benchmark PDEs for PINNs: Burgers' equation, Kirchhoff's plate bending equation and Helmholtz's equation. The results show that ReLoBRaLo is able to consistently outperform the baseline of existing scaling methods in terms of accuracy, while also inducing significantly less computational overhead.


\end{abstract}

\section{Introduction}
\label{intro}
The emergence of Physics-Informed Neural Networks (PINNs) \citep{Raissi2017} has sparked a lot of interest in domains that see themselves regularly confronted with problems in the low data regime. By leveraging well-known physical laws and incorporating them as implicit prior into the deep-learning pipeline, PINNs were shown to require little to no data in order to approximate partial differential equations (PDE) of varying complexity \citep{Raissi2017,hongweicollocationbendinganalysis}. 

We consider the case where PINNs are used for finding the unknown, underlying function proper to solve a parameterised PDE. PDEs generally consist of a governing equation and a set of boundary as well as initial conditions. When trained jointly, i.e. as multi-objective optimisation (MOO), these equations form a set of objective functions that guide the model to approximate a function that satisfies the PDE. While several established and well-studied numerical methods already exist for addressing this problem, such as the Finite Element Method (FEM), Finite Difference Method or Wavelets resp. Laplace transform methods \citep{bathe1996finite,hughes2012finite,smith1985a,Grossmann2007NumericalTO}, PINNs offer significant advantages, such as being end-to-end differentiable, mesh-free and avoiding the curse of dimensionality \citep{grohs2018proof,poggio2017and}. They could therefore prove useful in several engineering applications, such as inversion and surrogate modeling in solid mechanics \citep{haghighat2021physics}, design optimisation \citep{martins_ning_2021} or structural health monitoring and system identification \citep{yuan2020machine}. PINNs have also been successfully applied in computational fluid mechanics and dynamics for surrogate modelling of numerically expensive fluid flow simulations \citep{xiang2021self}, identification of hidden quantities of interest (velocity, pressure) from spatio-temporal visualisations of a passive scaler (dye or smoke) \citep{raissi2020hidden} or in an inverse heat transfer application setting in flow past a cylinder without thermal boundaries \citep{cai2020heat}.

However, further research is necessary to tackle current failure modes \citep{wang2020,mcclenny2020a}, one of which is the issue of gradient pathologies arising from imbalanced loss terms during training \citep{wanggradientpathologies}. With the various terms in the objective function stemming from physical laws, they are naturally bound to units of measurements that can vary significantly in magnitude. Consequently, the signal strengths of backpropagated gradients might differ from term to term and lead to pathologies that were shown to impede proper training and cause imbalanced solutions \citep{wanggradientpathologies}, hence posing challenges to global optimisation methods such as Adam, Stochastic Gradient Descent (SGD), or L-BFGS \citep{zhang1801a,adam,theodoridis2015a,kendall2018a}. As a counter-measure, every individual term $i$ may be scaled by a factor $\lambda_i$ in order to balance its contribution to the total gradient. However, manual tuning of these scaling factors requires laborious grid search and becomes intractable as the number of terms grows.

This work investigates different schemes aiming at adaptively balancing the contributions of multiple terms and their gradients in the loss function by selecting the optimal scaling factors $\lambda_i$ in order to improve approximation capabilities of PINNs. To this end, we compare the effectiveness of Learning Rate Annealing (LRAnnealing) \citep{wanggradientpathologies}, proposed in the context of PINNs, to two approaches originating from Computer Vision applications: GradNorm \citep{gradnorm} and SoftAdapt \citep{softadapt}. In addition, we derive and present our own variation of an adaptive loss scaling technique, \emph{ReLoBRaLo} (Relative Loss Balancing with Random Lookback), that we found to be more effective at similar efficiency compared to state-of-the-art by testing the algorithms on various benchmark problems for PINNs in the forward and inverse setting: Helmholtz, Burgers and Kirchhoff PDEs.

This paper is organised as follows: we first provide a short introduction to the problem as well as the state-of-the-art of PINNs in sec.~\ref{pinn}. Further methodical background on multi-objective optimisation (MOO), the framework of Physics-Informed Neural Networks (PINNs) and loss balancing for PINNs training is presented in sec.~\ref{sec:adaptive_loss_balancing}. In sec.~\ref{sec:ReLoBRaLo} we introduce ReLoBRaLo as a novel self-adaptive loss balancing method. Sec.~\ref{sec:results} reports numerical results of the developed approach against state-of-the-art methods for several examples in the forward and inverse setting: Burgers, Kirchhoff and Helmholtz PDEs. Sec.~\ref{sec:Ablation} presents results of ablation studies as well as a discussion of findings and drawing further conclusions on ReLoBRaLo and its hyperparameter settings across all examples of this paper. Finally, a summary together with an outlook is given in sec.~\ref{sec:summary}. All code produced within this publication is freely available and open access here: \href{https://github.com/rbischof/relative_balancing}{https://github.com/rbischof/relative\_balancing}.

\section{State-of-the-Art and Related Work}    \label{sec:SoA}

Using neural networks to approximate the solutions of Ordinary Differential Equations (ODEs) and PDEs has been the subject of several studies over the past decade. Initially, Lagaris et al. trained neural networks to solve ODEs and PDEs on a predefined set of grid points \citep{lagaris1998artificial,lagaris_2000}, while Sirignano proposed a method for solving high-dimensional PDEs through approximation of the solution by a neural network and especially emphasising training efficiency by incorporating mini-batch sampling in high dimensional settings compared to the computationally intractable finite mesh-based schemes \citep{Sirignano2018}. Raissi et al. introduced the term Physics-Informed Neural Networks (PINN) and provided empirical justification by numerical simulations for a variety of nonlinear PDEs, including the Navier–Stokes equation and the Burgers equation \citep{Raissi2018}. Shin et al. provided first theoretical justification for PINNs by demonstrating convergence of linear elliptic and parabolic PDEs in the L2 sense \citep{shin_2020}.

However, PINN training efficiency, convergence, and accuracy remain serious challenges \citep{raissi2019physics,xiang2021self}. Current research may be ordered into four main approaches: modifying structure of the NN, divide-and-conquer/domain decomposition, parameter initialisation and loss balancing.

Jagtap et al. adapted the typical NN architecture to PINNs by introducing parameters that scale the input to the activation functions and get updated alongside the network's parameters $\theta$ through gradient descent \citep{jagtap2020b,jagtap2020a}. The authors showed that the adaptive activation function significantly accelerated convergence and also improved solution accuracy. Kim et al. presented a fast and accurate PINN ROM with a nonlinear manifold solution representation, where the NN structure included an encoder and a decoder part \citep{kim2020}. Furthermore, a shallow masked encoder was trained using data from the full-order model simulations in order to use the trained decoder as representation of the nonlinear manifold solution. Peng et al. proposed dictionary-based PINNs to store and retrieve features and speed up convergence by merging prior information into the structure of NNs \citep{peng2020}. 

Other research focused on decomposing the computational domain in order to accelerate convergence. Jagtap et al. proposed conservative PINNs and extended PINNs that decompose the computational domain into several discrete sub-domains, each one solved independently using a separate, shallow PINN \citep{jagtap2020c,jagtap2020d}. Inspired by this work, Shukla et al. derived and investigated a distributed training framework for PINNs that used domain decomposition methods in both space and time-space \citep{shukla2021}. To accelerate convergence, the distributed framework combined the benefits of conservative and extended PINNs. The time-space domain may become very large when solving PDEs with long time integration, causing the training cost of NNs to become extremely expensive. To that end, Meng et al. proposed a parareal PINN to address the long-standing issue \citep{meng2020}. The authors decomposed the long-time domain into many discrete short-time domains using a fast coarse-grained solver. Training multiple PINNs with many small data sets was much faster than training a single PINN with a large data set. For PDEs with long-time integration, the parareal PINN achieved a significant speedup. Kharazmi et al. introduced hp-variational PINNs to divide the computational space into the trial space and test space by combining domain decomposition and projection onto high-order polynomials \citep{kharazmi2021}. A soft split of the problem domain incorporating variants of the Mixture-of-Experts approach were investigated by \citep{bischof2022mixture}.

In most works, researchers resort to the Xavier initialisation \citep{glorot_2010} for selecting the PINN's initial weights and biases. The effects of using more refined initialisation procedures has recently been gaining attention, with Liu et al. showing that a good initialisation can provide PINNs with a head start, allowing them to achieve fast convergence and improved accuracy \citep{liu2021novel}. Transfer learning for PINNs was introduced by Chakraborty et al. \citep{CHAKRABORTY2021} and Goswami et al. \citep{GOSWAMI2020} to initialise PINNs for dealing with multi-fidelity problems and brittle fracture problems, respectively. After their success in other fields of Deep Learning, meta-learning algorithms have also been implemented in the context of PINNs \citep{Rajeswaran2019,Smith2009,Finn2017a,Finn2018}, with Model-Agnostic Meta-Learning (MAML) being amongst the most popular ones \citep{Finn2017b}. Its second-order objective is to find an initialisation that is sub-optimal in itself, but from where the network requires only few labeled training samples and optimisation steps in order to specialise on a task and achieve high accuracy (few-shot learning). Subsequently, Nichol et al. proposed the REPTILE algorithm, which turns the second-order optimisation of MAML into a first-order approximation and therefore requires significantly less computation and memory while achieving similar performance \citep{Nichol2018}. Liu et al. applied the REPTILE algorithm to PINNs by regarding modifications of PDE parameters as separate tasks \citep{liu2021novel}. The resulting initialisation is such that the PINN converges in just a few optimisation steps for any choice of PDE parameters.

Using derivative information of the target function during training of a neural network was introduced by Czarnecki et al. under the term \emph{Sobolev Training} \citep{Czarnecki2017}. Sobolev Training proved to be more efficient in many applicable fields due to lower sample complexity compared to regular training. Son et al. enhance the concept of Sobolev Training in the strict mathematical sense using Sobolev norms in loss functions of neural networks for solving PDEs \citep{son2021sobolev}. It was found that these novel Sobolev loss functions lead to significantly faster convergence on investigated examples compared to traditional L2 loss functions. NNs were used in plain as well as a Sobolev Training manner for constitutive modelling, where it was shown that mechanical relations can be seen as Sobolev training to successfully encapsulate several aspects of the constitutive behavior, such as strain-stress-relationships arising from derivatives of a Helmholtz potential in hyperelasticity \citep{vlassis2021sobolev,vlassis2020geometric,krausphdthesis,kraus2020artificial}.

Colby et al. observed that a weighted scalarisation of the multiple loss functions, defined by the sampled data and physical laws for PINNs training, plays a significant role for convergence \citep{colby2020}. Wang et al. recently published a Learning Rate Annealing algorithm that employs back-propagated gradient statistics in the training procedure in order to adaptively balance the terms' contributions to the final loss \citep{wanggradientpathologies} and investigated the issue of vanishing and exploding gradients that currently limits the applicability of PINNs  \citep{wang2020}. To that end, the authors introduced a Neural Tangent Kernel (NTK), which appropriately assigns weights to each loss term at subtle performance improvement, in order to comprehend the training process for PINNs. Shin et al. developed the Lipschitz regularised loss for solving linear second-order elliptic and parabolic type PDEs \citep{shin_2020}. McClenny et al. proposed a method for updating the adaptation weights in the loss function in relation to network parameters \citep{mcclenny2020a}. 

\section{Physics-Informed Neural Networks (PINNs)}  \label{pinn}
This section reviews basic Physics-Informed Neural Networks (PINNs) concepts and recent developments.

Consider the following abstract parameterised and nonlinear PDE problem:
\begin{gather}  \label{eq:PDEgeneral}
\begin{aligned}
	\text{PDE}: \; & \mathcal{F} \left (\hat{\mathbf{u}}, \frac{\partial \hat{\mathbf{u}}}{\partial t}, \frac{\partial \hat{\mathbf{u}}}{\partial \mathbf{x}}, \cdots; \mathbf{\mu} \right) = 0, \quad \mathbf{x} \in \Omega, \; t \in \Upsilon \\
	\text{B.C.}: \; & \mathcal{B} \left (\hat{\mathbf{u}}, \frac{\partial \hat{\mathbf{u}}}{\partial \mathbf{x}}, \frac{\partial^2 \hat{\mathbf{u}}}{\partial \mathbf{x}^2}, \cdots \right) = 0, \quad \mathbf{x} \in \Gamma \\
	\text{I.C.}: \; & \mathcal{C} \left (\hat{\mathbf{u}}, \frac{\partial \hat{\mathbf{u}}}{\partial t}, \frac{\partial^2 \hat{\mathbf{u}}}{\partial t^2}, \cdots \right) = 0, \quad t \in \Upsilon
	\end{aligned}
\end{gather}
where $\mathbf{x} \in \mathbb{R}^d$ is the spatial coordinate and $t$ is the time; $\mathcal{F}$ denotes the residual of the PDE, containing the differential operators (i.e. $\partial_\mathbf{x}\hat{\mathbf{u}},\partial_t\hat{\mathbf{u}},...$); $\mathbf{\mu} = [\mu_1,\mu_2,...]$ are the PDE parameters; $\hat{\mathbf{u}}(\mathbf{x},t)$ is the solution of the PDE with initial condition $\mathcal{C}$ and boundary condition $\mathcal{B}$ (which can be Dirichlet, Neumann or mixed); $\Omega$, $\Gamma$ and $\Upsilon$ represent the spatial domain resp. boundary. A special example considered in this paper is the Burgers equation (given in Eq.~\ref{eq:burgers}): $\partial_t\mathbf{\hat{u}} + \mathbf{\hat{u}} \partial_{\mathbf{x}} \mathbf{\hat{u}} - \nu \partial^2_{\mathbf{x}} \mathbf{\hat{u}} = 0$ with PDE parameter $\mu$ as viscosity coefficient $\nu$. This paper is concerned with solving forward as well as inverse problems from different fields of application. For the forward problem, solutions of PDEs are to be inferred with fixed parameters $\mu$, while for the inverse problem setting, $\mu$ is unknown and has to be learned from observed data together with the PDE solution.

\begin{figure*}
    \centering
    \includegraphics[width=0.8\linewidth]{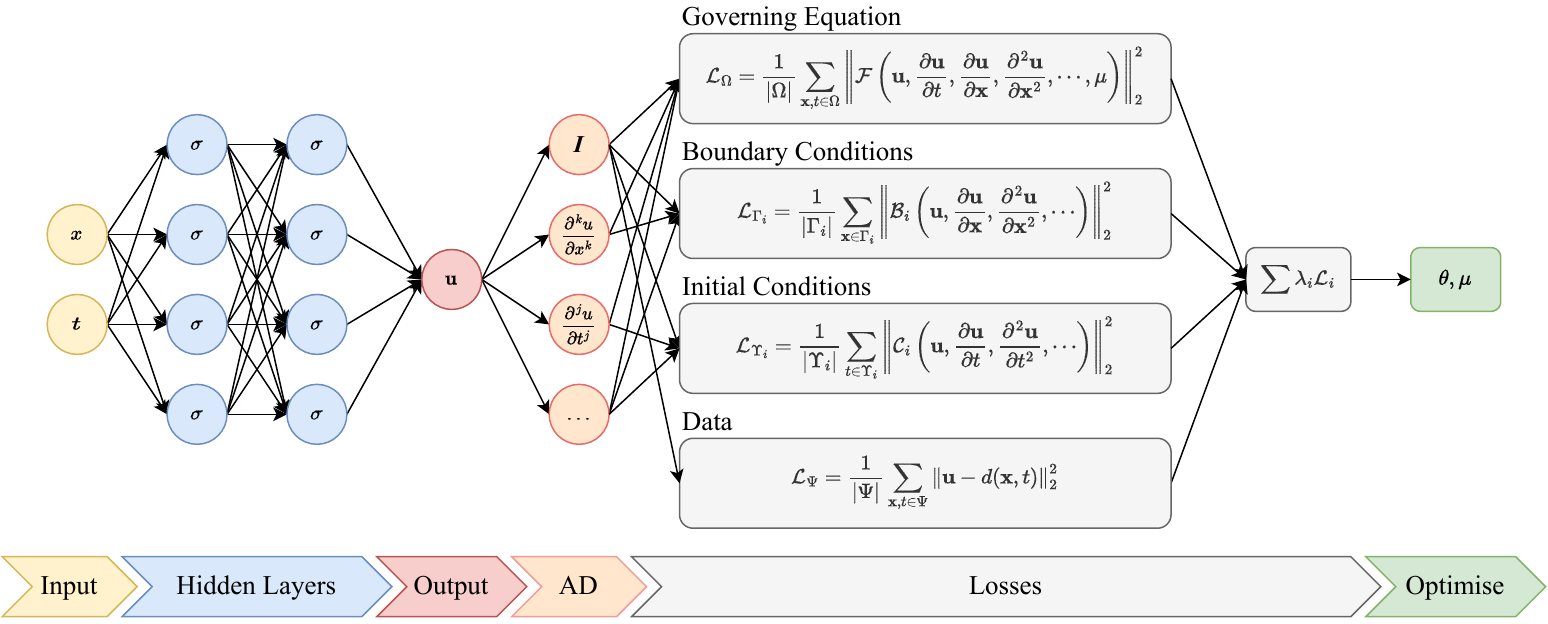}
    \caption{Schematic of a Physics-Informed Neural Network (PINN): A fully-connected feed-forward neural network with space and time coordinates $(\mathbf{x},t)$ as inputs, approximating a solution $\hat{\mathbf{u}}(\mathbf{x},t)$. Derivatives of $\mathbf{u}$ w.r.t. inputs are computed by automatic differentiation (AD) and then incorporated into residuals of the governing equations as the loss function, which is composed of multiple terms weighted by different coefficients. Parameters of the FCNN $\mathbf{\theta}$ and the unknown PDE parameters $\mathbf{\mu}$ may be optimised simultaneously by minimising the loss function.}
    \label{fig:pinnsGeneral}
\end{figure*}

Following the "vanilla" implementation of PINNs \citep{Raissi2017}, a fully-connected feed-forward neural network (FCNN) $U(\mathbf{x},t; \mathbf{\theta})$ is used to approximate the function $\hat{\mathbf{u}}(\mathbf{x},t)$ which solves the PDE. A FCNN consists of multiple hidden layers with trainable parameters (weights and biases; denoted by $\mathbf{\theta}$) and takes as inputs the space and time coordinates $(\mathbf{x},t)$, cf. fig.~\ref{fig:pinnsGeneral}. The losses are then defined as follows:
\begin{equation} 
	\centering
	\begin{aligned}
	    & \mathcal{L}_\Omega\, = \frac{1}{\vert \hat{\Omega} \vert} \sum_{\mathbf{x},t \in \hat{\Omega}} \left\lVert \mathcal{F} \left (\mathbf{u}, \frac{\partial \mathbf{u}}{\partial t}, \frac{\partial \mathbf{u}}{\partial \mathbf{x}}, \frac{\partial^2 \mathbf{u}}{\partial \mathbf{x}^2}, \cdots, \mathbf{\mu} \right) \right\rVert_2^2 \cr
	    & \mathcal{L}_{\Gamma_i} = \frac{1}{\vert \hat{\Gamma}_i \vert} \sum_{\mathbf{x}, t \in \hat{\Gamma}_i}{ \left\lVert \mathcal{B}_i \left (\mathbf{u}, \frac{\partial \mathbf{u}}{\partial \mathbf{x}}, \frac{\partial^2 \mathbf{u}}{\partial \mathbf{x}^2}, \cdots \right) \right\rVert_2^2 }\\
	    & \mathcal{L}_{\Upsilon_i} = \frac{1}{\vert \hat{\Upsilon}_i \vert} \sum_{\mathbf{x}, t \in \hat{\Upsilon}_i}{ \left\lVert \mathcal{C}_i \left (\mathbf{u}, \frac{\partial \mathbf{u}}{\partial t}, \frac{\partial^2 \mathbf{u}}{\partial t^2}, \cdots \right) \right \rVert_2^2 } \\ 
	    & \mathcal{L}_{\Psi}\, = \,\frac{1}{\vert \hat{\Psi} \vert}\, \sum_{\mathbf{x}, t \in \hat{\Psi}}\,{\left \lVert \mathbf{u} - d(\mathbf{x}, t) \right\rVert_2^2 }
    \end{aligned}
	\label{eq:PINN_PDE_Loss_def}
\end{equation}
where $\hat{\Omega}$ is a set of collocation points on the physical domain, $\hat{\Gamma}_i$ for the boundary conditions (BC), $\hat{\Upsilon}_i$ for the initial conditions (IC) and $\hat{\Psi}$ represents a set of measurements (data); the function $d$ maps $(\mathbf{x}, t)$ to measurements at those coordinates; $\mathbf{u}$ is the output from the neural network $U(\mathbf{x},t; \mathbf{\theta})$. PINNs are generally trained using the L2-norm (mean squared error / MSE) on uniformly sampled collocation points defined as a data set $\{\mathbf{x}_i,t_i\}_{i=1}^N$ prior to training. Note that the number of points $N$ (denoted by $|\cdot|$ in Eq.~\ref{eq:PINN_PDE_Loss_def}) may vary for different loss terms.

The objectives in Eq.~\ref{eq:PINN_PDE_Loss_def} are trained jointly and hence fall into the class of multi-objective optimisation (MOO) (cf. Eq.~\ref{eq:moo})

\begin{equation} 
	\centering
    \mathcal{L}(\mathbf{\mathbf{\theta, \mu}}) = (\mathcal{L}_\Omega, \mathcal{L}_{\Gamma_1},\dots,\mathcal{L}_{\Gamma_n}, \mathcal{L}_{\Upsilon_1},\dots, \mathcal{L}_{\Upsilon_m}, \mathcal{L}_{\Psi})^T
	\label{eq:PINN_MOO}
\end{equation}
where the individual terms can be interpreted in the following way:

\begin{itemize}
    \item the first term $\mathcal{L}_\Omega$ penalises the residual of the governing equations (PDEs), included in both the forward and inverse problem. 
    \item the following $n$ terms $\mathcal{L}_{\Gamma_i}$ enforce the boundary conditions (BCs), included only in the forward problem.
    \item the following $m$ terms $\mathcal{L}_{\Upsilon_i}$ enforce the initial conditions (ICs), included only in the forward problem.
    \item the last term $\mathcal{L}_{\Psi}$ makes the network approximate the measurements, included both in the forward (albeit not strictly necessary) and inverse problem.
\end{itemize}

A common approach for handling MOO is through linear scalarisation described in more detail in the following Section~\ref{sec:moo}.

\section{Multi-Objective Optimisation}  \label{sec:moo}
Multi-objective optimisation (MOO) is concerned with simultaneously optimising a set of $k > 1$, potentially conflicting objectives \citep{multitask_caruana,jones2002multi}.
\begin{equation} 
	\centering
	\mathcal{L}(\mathbf{\theta}) = (\mathcal{L}_1(\mathbf{\theta}), \dots, \mathcal{L}_k(\mathbf{\theta}))^T
    \label{eq:moo}
\end{equation}
which can be turned into a single objective through linear scalarisation:
\begin{equation} 
	\centering
	\mathcal{L}(\mathbf{\theta}) = \sum_{i = 1}^k \lambda_i \mathcal{L}_i(\mathbf{\theta}), \quad \lambda_i \in \mathbb{R}_{>0}
	\label{eq:linear_scalarisation}
\end{equation}

Many problems in engineering, natural sciences, or economics can be formulated as multi-objective optimisations and generally require trade-offs to simultaneously satisfy all objectives to a certain degree \citep{moo_engineering}. The solution of MOO models is usually expressed as a set of Pareto optima, representing these optimal trade-offs between given criteria according to the following definitions \citep{pareto_multi_task}:
\begin{definition}
    \label{pareto_dominance}
    A solution $\hat{\mathbf{\theta}} \in \Omega$ Pareto dominates solution $\mathbf{\theta}$ (denoted $\hat{\mathbf{\theta}} \prec \mathbf{\theta}$) if and only if $\mathcal{L}_i(\hat{\mathbf{\theta}}) \leq \mathcal{L}_i(\mathbf{\theta}), \forall i \in \{1, \dots, m\}$ and $\exists j \in \{1, \dots, m\}$ such that $\mathcal{L}_j(\hat{\mathbf{\theta}}) < \mathcal{L}_j(\mathbf{\theta})$.
\end{definition}
\begin{definition}
    \label{pareto_optimal}
    A solution $\hat{\mathbf{\theta}} \in \Omega$ is said to be Pareto optimal if $\forall {\mathbf{\theta}} \in \Omega, \hat{\mathbf{\theta}} \preceq \mathbf{\theta}$. The set of all Pareto optimal points is called the Pareto set and the image of the Pareto set in the loss space is called the Pareto front.
\end{definition}

In theory, a Pareto optimal solution $\mathbf{\theta}$ is independent of the scalarisation \citep{efficient_moo}. However, when using neural networks for MOO, the solution space becomes highly non-convex. Thus, although neural networks are universal function approximators \citep{Hornik1990}, they are not guaranteed to find the globally optimal solution through gradient-based optimisation. Scaling the loss space therefore provides the option of guiding the gradients into having an a priori deemed desirable property. However, manually finding optimal $\mathbf{\lambda_i}$ requires laborious grid search and becomes intractable as $k$ gets large. Furthermore, one might want to let $\mathbf{\lambda_i}$ evolve over time. This raises the need for an automated scheme to dynamically choose the scalings $\mathbf{\lambda_i}$.

\section{Adaptive Loss Balancing Schemes}  \label{sec:adaptive_loss_balancing}
This section reviews different methods aiming at balancing the various terms within multi-objective optimisation. To this end, we compare the effectiveness of Learning Rate Annealing \citep{wanggradientpathologies}, proposed in the context of PINNs as well as two approaches originating from Computer Vision applications: GradNorm \citep{gradnorm} and SoftAdapt \citep{softadapt}. This forms the basis for deriving and presenting our own loss balancing method as given in sec.~\ref{sec:ReLoBRaLo}.

\subsection{Learning Rate Annealing}
\label{sec:lrannealing}
Wang et al. \cite{wanggradientpathologies} conducted a study on gradients in PINNs and identified pathologies that explained some failure modes. One pathology is gradient stiffness in the boundary conditions caused by the imbalance amongst the different loss terms. As a remedy, it is proposed to adaptively scale the loss using gradient statistics, thus reducing the laborious tuning of these hyperparameters. 

\begin{equation}
    \label{eq:lr_annealing}
    \begin{aligned}
        &\hat{\lambda}_i(t) = \frac{\text{max}\{\vert \nabla_\mathbf{\theta} \mathcal{L}_\Omega(t) \vert\}}{\overline{\vert \nabla_\mathbf{\theta} \mathcal{L}_{\{\Gamma, \Upsilon\}_i}(t) \vert}}, \text{ } i \in \{1, \dots, k\} \\
        &\lambda_i(t) = \alpha\lambda_i(t-1) + (1-\alpha)\hat{\lambda}_i(t) \\
        &\mathbf{\theta}^{(t+1)} = \mathbf{\theta}^{(t)} - \eta\nabla_\mathbf{\theta} \left(\mathcal{L}_\Omega(t) + \sum_{i=1}^k \lambda_i(t) \mathcal{L}_{\{\Gamma, \Upsilon\}_i}(t)\right)
    \end{aligned}
\end{equation}
where $\overline{\vert \nabla_\mathbf{\theta} \mathcal{L}_{\Gamma_i}(t) \vert}$ is the mean of the gradient w.r.t. the parameters $\mathbf{\theta}$; $\alpha$ is a hyperparameter with a value $\alpha = 0.9$ recommended by the authors.

With this method, whenever the maximum value of $\vert \nabla_\mathbf{\theta} \mathcal{L}_\Omega(t) \vert$ grows considerably larger than the average value in $\vert \nabla_\mathbf{\theta} \mathcal{L}_{\{\Gamma, \Upsilon\}}(t) \vert$, the scalings $\mathbf{\hat{\lambda}_i}(t)$ correct for this discrepancy such that all gradients have similar magnitudes. Additionally, exponential decay is used in order to smoothen the balancing and avoid drastic changes of the loss space between optimisation steps.

This procedure induces a few drawbacks. Its unboundedness potentially involves up- or down-scaling of terms by means of several orders of magnitude. The up-scaling in particular can cause problems similar to the effect of choosing a learning rate that is too large and therefore leads to repeatedly overshooting the objective. Furthermore, scaling all terms to have the same magnitude throughout training can incite the network to optimise for the "low-hanging fruit". A term, whose loss decreased considerably in the last optimisation step, will see its contribution to the total gradient scaled back up to the same magnitude to match the other terms. Therefore, the network might focus on the objectives that are easiest to optimise for.

\subsection{GradNorm} \label{sec:gradnorm}
Chen et al. \citep{gradnorm} take a different approach and make the scalings $\mathbf{\lambda}_i$ trainable. The updates on these trainable scalings are chosen such that all terms improve at the same relative rate w.r.t. their initial loss and performed by a separate optimiser. A term that improved at a higher rate since the beginning of training compared to the other terms, gets a weaker scaling until all terms have made the same relative progress. Therefore, one could argue that they weakly enforce each optimisation step to Pareto dominate (cf. definition \ref{pareto_dominance}) its predecessor. The loss for updating the scalings within GradNorm is computed as follows:

\begin{equation}
    \label{eq:granorm_scalings_loss}
    \begin{aligned}
        \mathcal{L}(t; \mathbf{\lambda}) = \sum_{i = 1}^k \left\vert G_\mathbf{\theta}^{(i)}(t) - \overline{G}_\mathbf{\theta}(t) \times [r_i(t)]^\alpha \right\vert_1
    \end{aligned}
\end{equation}
where $G_\mathbf{\theta}^{(i)}(t) = \lVert \nabla_\mathbf{\theta} \lambda_i \mathcal{L}_i(t) \rVert_2$ is the $L_2$ norm of the gradient w.r.t. the network parameters $\mathbf{\theta}$ for the scaled loss of objective $i \in \{1, \dots, k\}$; $\overline{G}_\mathbf{\theta}(t) = \frac{1}{k} \sum_{i=1}^k G_\mathbf{\theta}^{(i)}(t)$ is the average of all gradient norms; $r_i(t) = \mathcal{L}_i(t) / (\mathcal{L}_i(0) \cdot \overline{\mathcal{L}}(t))$ defines the rate at which term $i$ improved so far; $\alpha$ is a hyperparameter representing the strength of the restoring force which pulls tasks back to a common training rate. Note that $\overline{G}_\mathbf{\theta}(t) \times [r_i(t)]^\alpha$ is the desirable value that $G_\mathbf{\theta}^{(i)}(t)$ should take on, so gradients must be prevented from flowing through this expression. The final loss for updating the networks parameters is then simply a linear scalarisation with the scalings that were previously updated:

\begin{equation}
    \label{eq:gradnorm_params_loss}
    \begin{aligned}
        \mathcal{L}(t; \mathbf{\theta}) = \sum_{i = 1}^k \lambda_i(t) \mathcal{L}_i(t)
    \end{aligned}
\end{equation}

This algorithm is fairly evolved and, despite solving some of Learning Rate Annealing's issues, it still requires a separate backward-pass for each task, which becomes prohibitively expensive as $k$ gets large. Furthermore, it relies on two separate optimisation rounds at each step: one for adapting the scalings $\mathbf{\lambda}_i$ and another for updating the weights $\mathbf{\theta}$. By means of Eq.~\ref{eq:moo}, GradNorm can thus be formulated as a scalarised MOO via:

\begin{equation}
    \label{eq:gradnorm_moo}
    \begin{aligned}
        \mathcal{L}(t) = (\mathcal{L}(t; \mathbf{\theta}), \mathcal{L}(t; \lambda))^T
    \end{aligned}
\end{equation}
which in turn requires empirical hyperparameter tuning (learning rate, initialisation, etc.) to keep the system balanced - exactly the problem we are actually trying to solve through the use of adaptive loss balancing schemes.

\subsection{SoftAdapt} \label{sec:relative}
Similar to GradNorm, SoftAdapt \citep{softadapt} leverages the ansatz of relative progress in order to balance the loss terms. However, the authors relax it by only considering the previous time-step $\mathcal{L}_i(t-1)$ and taking the difference between time steps instead of the division. The scalings are then normalised by using a softmax function:
\begin{equation}
    \label{eq:softadapt}
    \begin{aligned}
        \lambda_i(t) = \frac{\operatorname{exp}\left(\mathcal{T}(\mathcal{L}_i(t)-\mathcal{L}_i(t-1))\right)}{\sum_{j=1}^k \operatorname{exp}\left(\mathcal{T}(\mathcal{L}_j(t)-\mathcal{L}_j(t-1))\right)}, \text{ } i \in \{1, \dots, k\}
    \end{aligned}
\end{equation}
where $\mathcal{L}_i^{(t)}$ is the loss of term $i$ at optimisation step $t$.

SoftAdapt also differs from GradNorm in the sense that it does not require gradient statistics and thus eliminates the need of performing separate backward passes for each objective. Instead, it makes use of the fact that magnitudes in the gradients directly depend on the magnitudes of the terms in the loss function and therefore aims at achieving the balance solely through loss statistics. This is obviously true only if the same loss function is used for every objective (e.g. the $L_2$ loss). However, this setting generalises to a vast majority of applications involving PINNs.

\section{\textbf{Re}lative \textbf{Lo}ss \textbf{Ba}lancing with Random \textbf{Lo}okback (ReLoBRaLo)}  \label{sec:ReLoBRaLo}
Drawing inspiration from existing balancing techniques as outlined in sec.~\ref{sec:adaptive_loss_balancing}, we propose a novel method and implementation for balancing the multiple terms in the scalarised MOO loss function for training of PINNs upon:
\begin{itemize}
    \item SoftAdapt's concept of operating on loss statistics as opposed to gradient statistics is employed. A computationally inexpensive softmax ensures the sum of scalings is bounded.
    \item Inspired by GradNorm, the progress is calculated by dividing the loss at the current iteration $\mathcal{L}_i(t)$ by the loss at the previous iteration $\mathcal{L}_i(t-1)$.
    \item Similarly to Learning Rate Annealing, the scalings are updated using an exponential decay in order to utilise loss statistics from more than just one training step in the past.
    \item In addition, a random lookback (called \emph{saudade} $\rho$) is introduced into the exponential decay, which decides whether to use the previous steps' loss statistics to compute the scalings, or whether to look all the way back until the start of training $\mathcal{L}_i^{(0)}$.
\end{itemize}

\begin{equation}
    \label{eq:relative_annealing}
    \begin{aligned}
        &\lambda_i^{\textit{bal}}(t, t') = m\cdot\frac{\operatorname{exp}\left(\frac{\mathcal{L}_i(t)}{\mathcal{T} \mathcal{L}_i(t')}\right)}{\sum_{j=1}^m \operatorname{exp} \left(\frac{\mathcal{L}_j(t)}{\mathcal{T} \mathcal{L}_j(t')} \right)}, \; i \in \{1, \dots, m\}\\
        &\lambda_{i}^{\textit{hist}}(t) = \rho\lambda_i(t-1) + (1-\rho)\lambda_i^{\textit{bal}}(t, 0))\\
        &\lambda_i(t) = \alpha\lambda_{i}^{\textit{hist}} + (1-\alpha)\lambda_i^{\textit{bal}}(t, t-1)
    \end{aligned}
\end{equation}
where $\alpha$ is the exponential decay rate, $\rho$ is a Bernoulli random variable and $\mathbb{E}[\rho]$ should be chosen close to 1. The intermediate step $\lambda_i^{\textit{bal}}(t, t')$ calculates scalings based on the relative improvements of each term between time steps $t'$ and $t$. The following step $\lambda_{i}^{\textit{hist}}(t)$ defines, whether the scalings calculated in the previous time step ($\rho$ evaluates to 1) or the relative improvements since the beginning of training ($\rho$ evaluates to 0) should be carried forward. Note that this concept of randomly retaining or discarding the history of scalings is what we denote as "random lookbacks". Finally, the scaling $\lambda_i(t)$ for term $i$ is obtained by means of an exponential decay, where $\alpha$ controls the weight given to past scalings versus the scalings calculated in the current time step.

This method is an attempt at combining the best attributes of the aforementioned approaches into a new scheme for scalarised MOO objective functions. First and foremost, it still weakly enforces every training step to Pareto dominate its predecessor, which is an important property in physical applications. It also avoids using gradient statistics, making it considerably more efficient than Learning Rate Annealing and GradNorm. Furthermore, it reduces drastic changes in the loss space by using exponential decay and can easily be adapted to use more or fewer information of past optimisation steps by tuning the hyperparameter $\alpha$. One can think of $\alpha$ as the model's ability to remember the past, with a high alpha giving lots of weight to past loss statistics, while a lower alpha increases stochasticity. Setting $\alpha = 1$ results in each term's relative progress being computed w.r.t. the initial loss $\mathcal{L}_i^{(0)}$. However, we found this to be too restrictive, since it causes the model to stop making progress as soon as one term reaches a local minimum. We chose values $\alpha$ between 0.9 and 0.999 and report the effects of varying this hyperparameter in sec.~\ref{sec:Ablation}.

Choosing the value of $\alpha$ also requires to make a trade-off: a high value means the model will remember potential deterioration of certain terms for longer and therefore leave a longer time frame in order to compensate them. However, it also induces a latency between a term starting to deteriorate and the scalings $\lambda_i$ reacting accordingly. We therefore study the effect of introducing the saudade Bernoulli random variable $\rho$ that causes the model to occasionally look back until the start of training. $\mathbb{E}[\rho] = 0$ is maximum saudade as it always takes the loss value of the initial training step, while $\mathbb{E}[\rho] = 1$ corresponds to minimum saudade, taking only into account the last value from the history of the $i$-th scaling factor. Selecting $\mathbb{E}[\rho]$ somewhere between 0 and 1 allows to set a lower value for $\alpha$, thus making the model more flexible while still occasionally "reminding" it of the progress made since the start of training. Furthermore, the random lookback can give episodic new impulses and let the model escape local minima by changing the loss space, as well as inciting it to explore more of the parameter space. In case the impulse would turn out to have a negative effect on the accuracy, one can still choose to roll back and reset the network's parameters $\theta$ to the previous state.

The last hyperparameter is the so-called temperature $\mathcal{T}$. Setting $\mathcal{T} \xrightarrow{} \infty$ re-calibrates the softmax to output uniform values and thus all $\hat{\lambda}_i^{(t)} = 1$. On the other hand, $\mathcal{T} \xrightarrow{} 0$ essentially turns the softmax into an argmax function, with the scaling $\hat{\lambda}_i^{(t)} = k$ resulting for the term with the lowest relative progress and $\hat{\lambda}_i^{(t)} = 0$ for all others. A pedagogical example with interpretation of expected behaviours and how to draw conclusions from the histories of the scalings is given for Burgers' equation in sec.~\ref{sec:burgers}.

Note that the network should be prevented from optimising this expression. This can be achieved by stopping the gradients from flowing through the calculation of the scalings. Also, depending on the problem at hand, $\operatorname{exp}(\mathcal{L}_i(t)/(\mathcal{T} \mathcal{L}_i(t')))$ could evaluate to a very large number, thus leading to overflows. This issue can be preemptively tackled by subtracting a large number (e.g. $10^{-9}$) from the input to the softmax.

\section{Hyperparameter Tuning and Meta Learning}  \label{Hyperparametertuning}
This paper uses grid search in combination with Bayesian Optimisation (BO) \citep{gridsearch,bayesian_optimization_snoek,bayesian_opt_implementation} for hyperparameter tuning. This study uses hyperparameters for defining the NN architecture ($d_K$ hidden layers and $w_K$ neurons per layer) and training settings (learning rate $l_r$, exponential decay rate $\alpha$ and saudade $\rho$). Tab.~\ref{tab:bayes_opt_parameters} contains the ranges and distributions for the hyperparameters. Bayesian Optimisation reduces the empiricism of selecting the PINNs hyperparameters to learn an optimal NN structure. First, 20 random points in the hyperparameter space are sampled and evaluated. The model's performance at those points serves as evidence for fitting prior Gaussian Processes in order to estimate the unknown loss function w.r.t. the hyperparameters. Using Expected Improvement (EI) \citep{expected_improvement}, further 80 points are then sampled and evaluated to refine the prior. This procedure provides an educated guess as to which are the optimal hyperparameters for the task at hand. Finally, we fine-tune the results by performing fine-grained grid search around the hyperparameters returned by Bayesian Optimisation (BO).

\begin{table}[h]
  \centering
    \begin{tabular}{lcc}
    Hyperparameter      & Range & Log-scaling \\ 
    \midrule            
    Learning Rate $l_r$     & [$10^{-6}, 10^{-2}$] & yes\cr
    Layers $d_K$            & [2, 4] & no \cr
    Neurons per Layer  $w_K$           & [32, 512] & no \cr
    Exponential Decay Rate $\alpha$          & [0, 1] & no \cr
    Temperature $\mathcal{T} = 10^{t}$                 & [$10^{-6}, 10^{2}$] & yes \cr
    Expected Saudade $\mathbb{E}[\rho]$& [0, 1] & no \cr
    Activation function $\sigma$ & \{\textit{tanh}, \textit{sigmoid}\}& no
  \end{tabular}
    \caption{Hyperparameters for architecture and training settings together with ranges as used for Bayesian Optimisation.}
  \label{tab:bayes_opt_parameters}
\end{table}

Within this study, the exact same Bayesian Optimisation configuration was used for all examples presented in sec.~\ref{sec:results}, hence it is sufficient to only display tab.~\ref{tab:bayes_opt_parameters}. Respective results of the Grid Search and Bayesian Optimisation can also be found in sec.~\ref{sec:results}.

\section{Results} \label{sec:results}
We evaluate the different balancing schemes on three problems (Burgers equation, Kirchhoff plate bending and Helmholtz equation) originating from physics-informed deep learning, where the objective function consists of various terms of potentially considerably different magnitudes and compare their performances, as well as their computational efficiency. Training was done on networks of varying depth and width (acc. to tab.~\ref{tab:bayes_opt_parameters}) and limited to $10^5$ steps of gradient descent (GD) using the Adam optimiser \citep{adam}. Additionally, we reduced the learning rate by a multiplicative factor of 0.1 whenever the optimisation stopped making progress for over 3’000 optimisation steps and finally used early stopping in case of 9’000 steps without improvement. 
When addressing the inverse problem, i.e. approximating a set of measurements while subjecting the network to PDE constraints for finding unknown PDE parameters $\mu$, we further investigated the payoff of using two separate optimisers: one for updating network weights $\theta$, and a separate one for updating PDE parameters $\mu$.
Further details on hyperparameter tuning and meta learning is given in sec.~\ref{sec:Ablation}.

\subsection{Burgers' Equation}
\label{sec:burgers}
Burgers’ equation is a one-dimensional PDE describing the main properties of the Navier-Stokes equations \citep{Orlandi2000} used i.a. to model shock waves, gas dynamics, or traffic flow \citep{burgers_pde}. Using Dirichlet boundary conditions, the PDE takes the following form:
\begin{equation}
    \label{eq:burgers}
    \begin{aligned}
        &\frac{\partial u}{\partial t} + u \frac{\partial u}{\partial x} - \nu \frac{\partial^2 u}{\partial x^2} = 0, \quad x \in [-1, 1], \quad t \in [0, 1]\\
        &u(0, x) = - sin(\pi x)\\
        &u(t, -1) = u(t, 1) = 0
    \end{aligned}
\end{equation}

At first, we investigate the solution of the forward problem, where we set the PDE parameter $\mu := \nu = \frac{1}{100 \pi}$. In order to find the latent function $\hat{\mathbf{u}}(\mathbf{x},t)$, we can parameterise it with a neural network $U(x,t; \mathbf{\theta})$ and turn the set of equations into a linear scalarised objective (cf. Eq.~\ref{eq:linear_scalarisation}) of Mean Squared Errors (MSE). This loss function will weakly enforce the network to approximate the PDE solution $\hat{\mathbf{u}}(\mathbf{x},t)$.

\begin{equation}
    \label{eq:burgers_loss}
    \begin{array}{@{}lll@{}}
        \mathcal{L}_{\Omega} &=& \frac{1}{\vert \hat{\Omega} \vert}\sum_{(x,t) \in \hat{\Omega}}{ \left\lVert \frac{\partial U}{\partial t} + U\frac{\partial U}{\partial x} - \nu \frac{\partial^2 U}{\partial x^2} \right\rVert_2^2 }\\
        \mathcal{L}_{\Gamma_{1}} &=& \frac{1}{\vert \hat{\Gamma}_{1} \vert}\sum_{t \in \hat{\Gamma}_{1}} \left\lVert U(-1, t; \mathbf{\theta}) \right\rVert_2^2 \\
        \mathcal{L}_{\Gamma_{2}} &=& \frac{1}{\vert \hat{\Gamma}_{2} \vert}\sum_{t \in \hat{\Gamma}_{2}}\left\lVert U(1, t; \mathbf{\theta}) \right\rVert_2^2 \\
        \mathcal{L}_{\Upsilon} &=& \frac{1}{\vert \hat{\Upsilon} \vert} \sum_{x \in \hat{\Upsilon}}{ \left\lVert U(x,0; \mathbf{\theta}) + \sin(\pi x) \right\rVert_2^2 }
    \end{array}
\end{equation}

For the forward problem, PINNs training induces the following loss function employed during training:
\begin{equation}
    \label{eq:burgers_scalarised_loss}
    \begin{aligned}
        \mathcal{L} = \lambda_0 \mathcal{L}_\Omega + \lambda_1 \mathcal{L}_{\Gamma_{1}} + \lambda_2 \mathcal{L}_{\Gamma_{2}} + \lambda_3 \mathcal{L}_{\Upsilon}
    \end{aligned}
\end{equation}

\begin{figure*}
	\centering
	\begin{tabular}{c c c}
		\hline  
		\cellcolor{hellgrau} \small (a) FEM-Result  & \cellcolor{hellgrau} \small (b) PINN-Result & \cellcolor{hellgrau} \small (c) Squared Error \\ 
		\hline  
		\vspace{0.5mm}
		\includegraphics[width=0.25\linewidth]{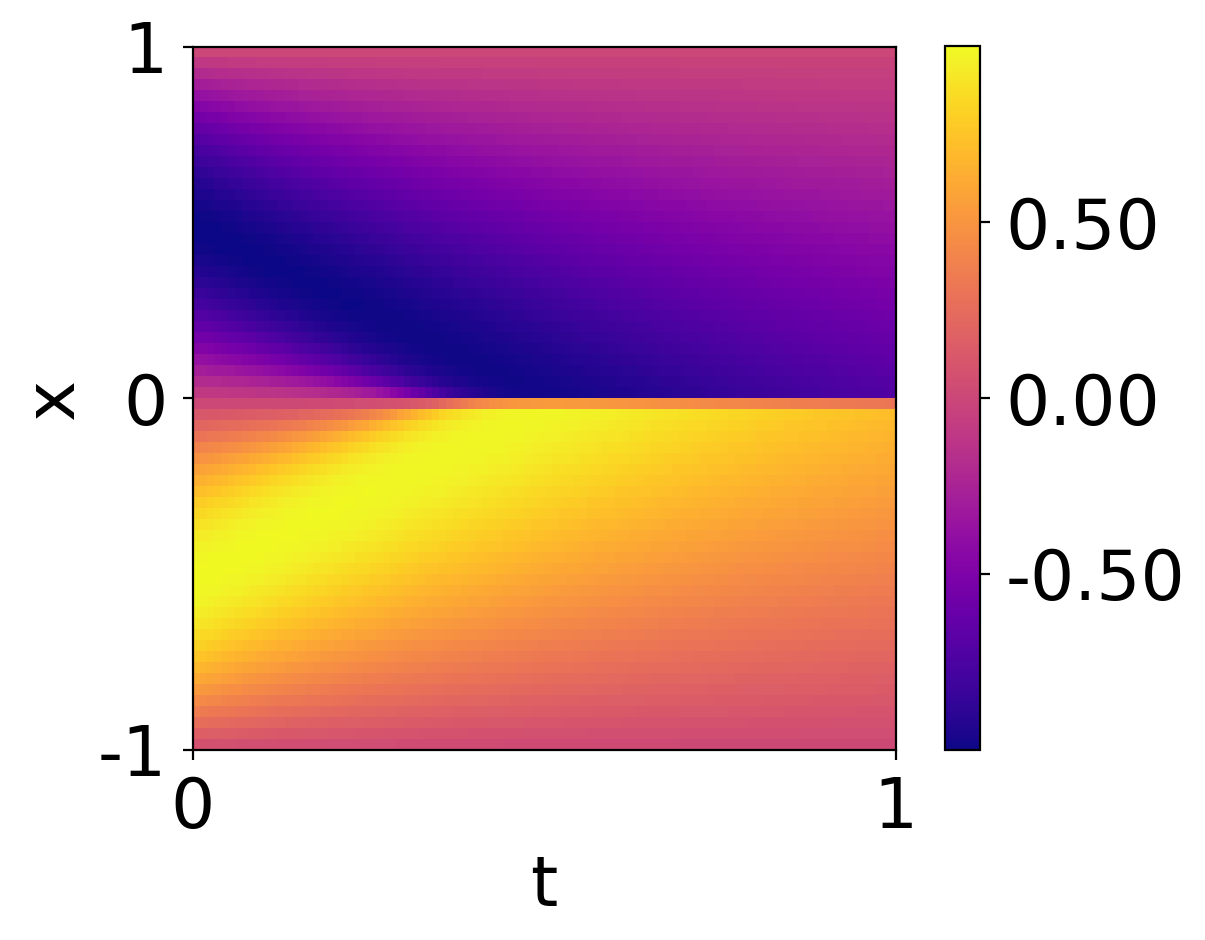} & \includegraphics[width=0.25\linewidth]{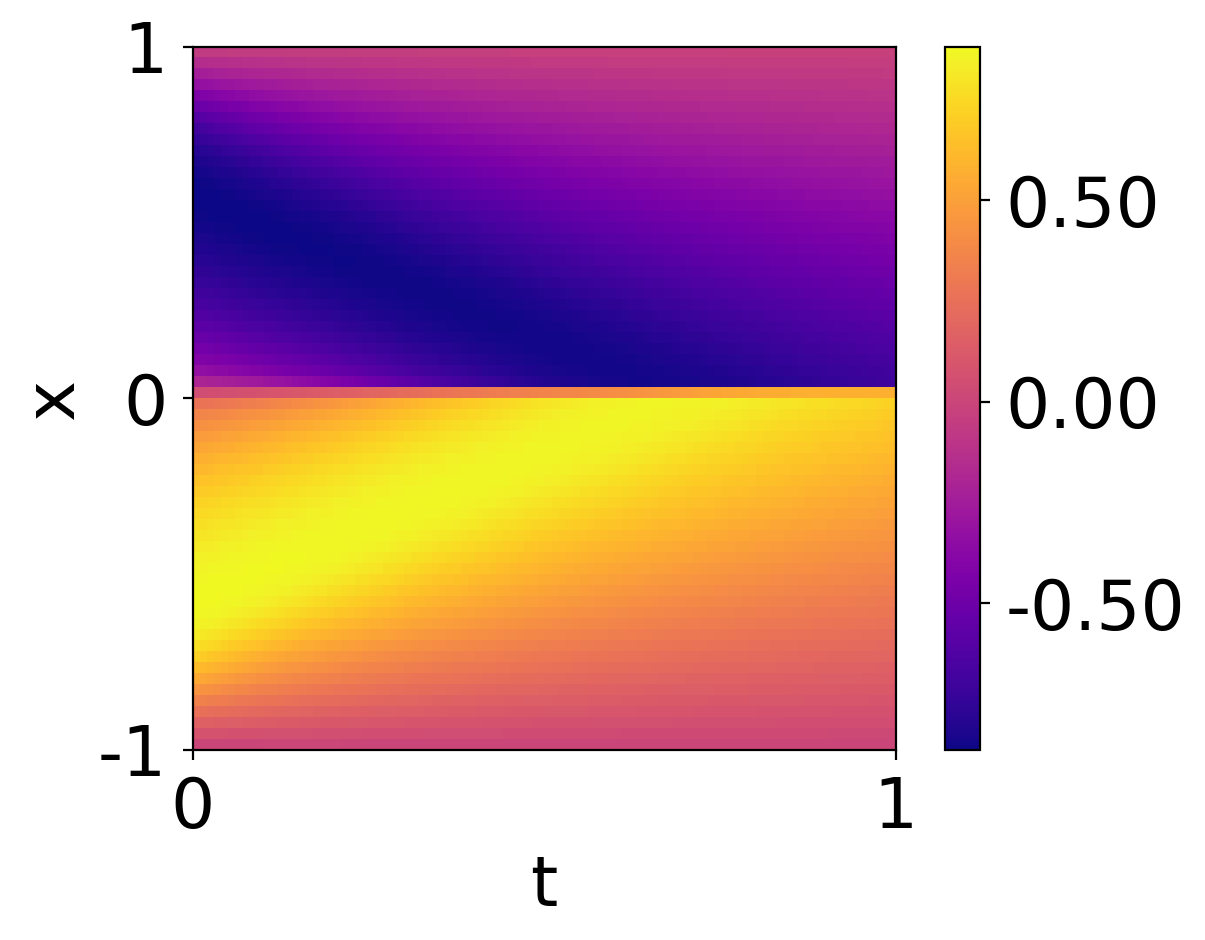}  & \includegraphics[width=0.255\linewidth]{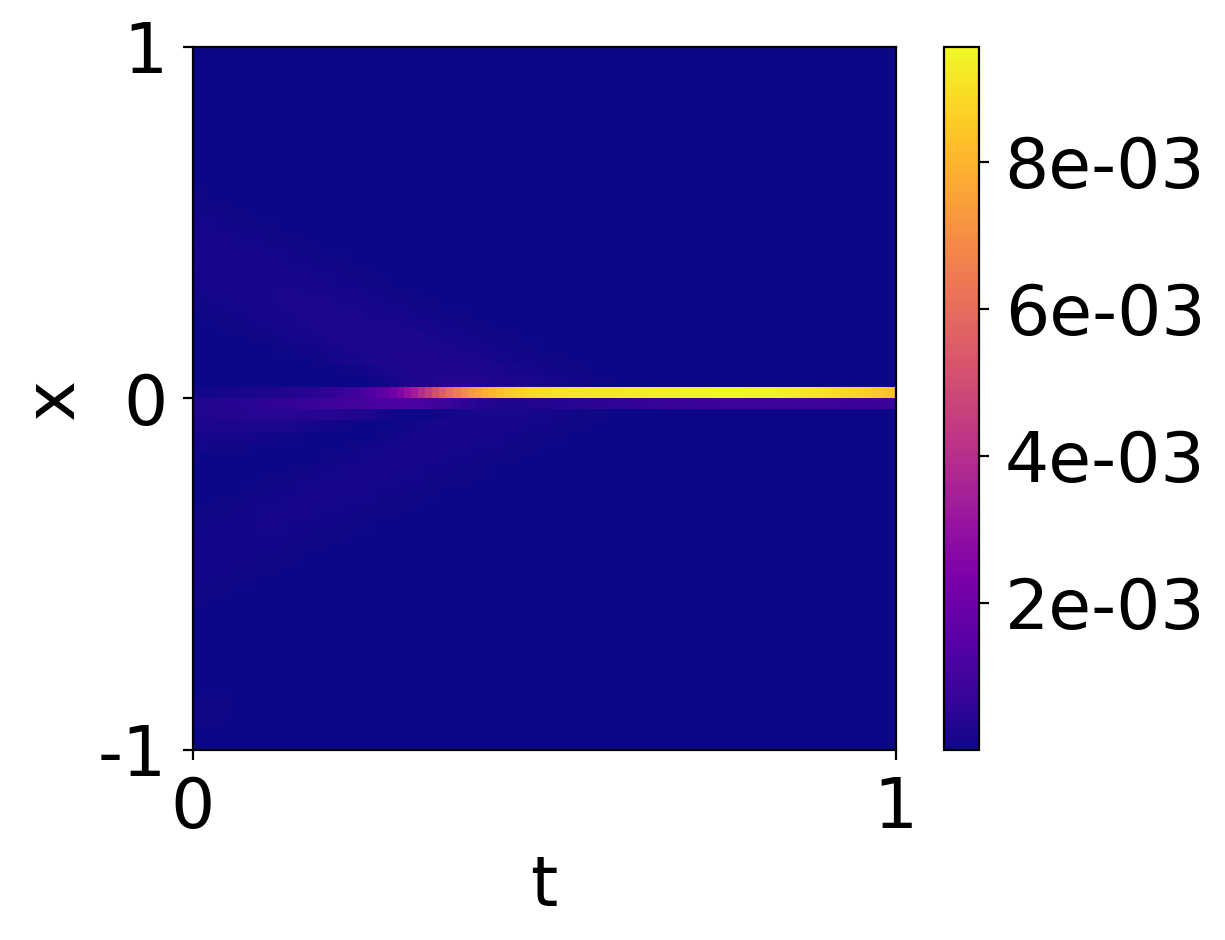} \\
	\end{tabular} 
    \caption{Burgers' equation problem: (a) FEM reference solution, (b) PINNs results predicted with a fully-connected network consisting of two layers and 128 nodes each, and (c) squared error.}
    \label{fig:pred_burgers}
\end{figure*}

After a successful convergence of the PINNs training using ReLoBRaLo, we obtain the results displayed in fig.~\ref{fig:pred_burgers}(b), whereas the final algorithm settings are reported in tab.~\ref{tab:bayes_opt_parameters_final}. As there is no analytical solution available for the Burgers equation, we compared the results to a reference solution calculated using the finite element method (FEM), displayed in fig.~\ref{fig:pred_burgers}(a). A plot of the squared difference in $u$ as given by the FEM and PINNs is shown in fig.~\ref{fig:pred_burgers}(c) and delivers a relative max error of below 5\%.

However, Burgers' equation can also be turned into an \textit{inverse problem} by regarding the PDE parameter $\nu$ as an unknown to be estimated from a set of observations (i.e. data) over the spatial and temporal domain. In this setting, the PINNs induced loss function to be deployed reads:
\begin{equation}
    \label{eq:burgers_inverse_loss}
    \begin{aligned}
        \mathcal{L} = \lambda_0 \frac{1}{\vert \hat{\Omega} \vert}&\sum_{(x,t) \in \hat{\Omega}}{ \left\lVert \frac{\partial U}{\partial t} + U\frac{\partial U}{\partial x} - \nu \frac{\partial^2 U}{\partial x^2} \right\rVert_2^2 }\\ +\; \lambda_1 \frac{1}{\vert \hat{\Psi} \vert}&\sum_{(x,t) \in \hat{\Psi}}{ \left\lVert U - u \right\rVert_2^2 } 
    \end{aligned}
\end{equation}

Similar to the network's weights and biases, the additional trainable PDE variable $\mu$ (here viscosity $\nu$) is now also updated through gradient descent:
\begin{equation}
    \label{eq:burgers_inverse_gradient_descent}
    \begin{aligned}
        &\mathbf{\theta}^{(t)} = \mathbf{\theta}^{(t-1)} - \eta \nabla_\mathbf{\theta} \mathcal{L}\\
        &\mu^{(t)} = \mu^{(t-1)} - \eta \nabla_\mu \mathcal{L}\\
    \end{aligned}
\end{equation}

Measurement data $\Psi$ for the inverse problem setting were obtained from our reference solution computed using the FEM without addition of noise. At every iteration, we sample from the available data in order to generate a batch of collocation points.

\begin{figure}
    \centering
	\hspace{3mm}\begin{tabular}{c c}
	\hline 
	\cellcolor{hellgrau} \small (i) $L_2$ Convergence  & \cellcolor{hellgrau} \small (ii) ReLoBRaLo Scaling \\ 
	\hline 
	\end{tabular}
    \begin{subfigure}{\linewidth}
    \centering
        \includegraphics[width=0.45\linewidth]{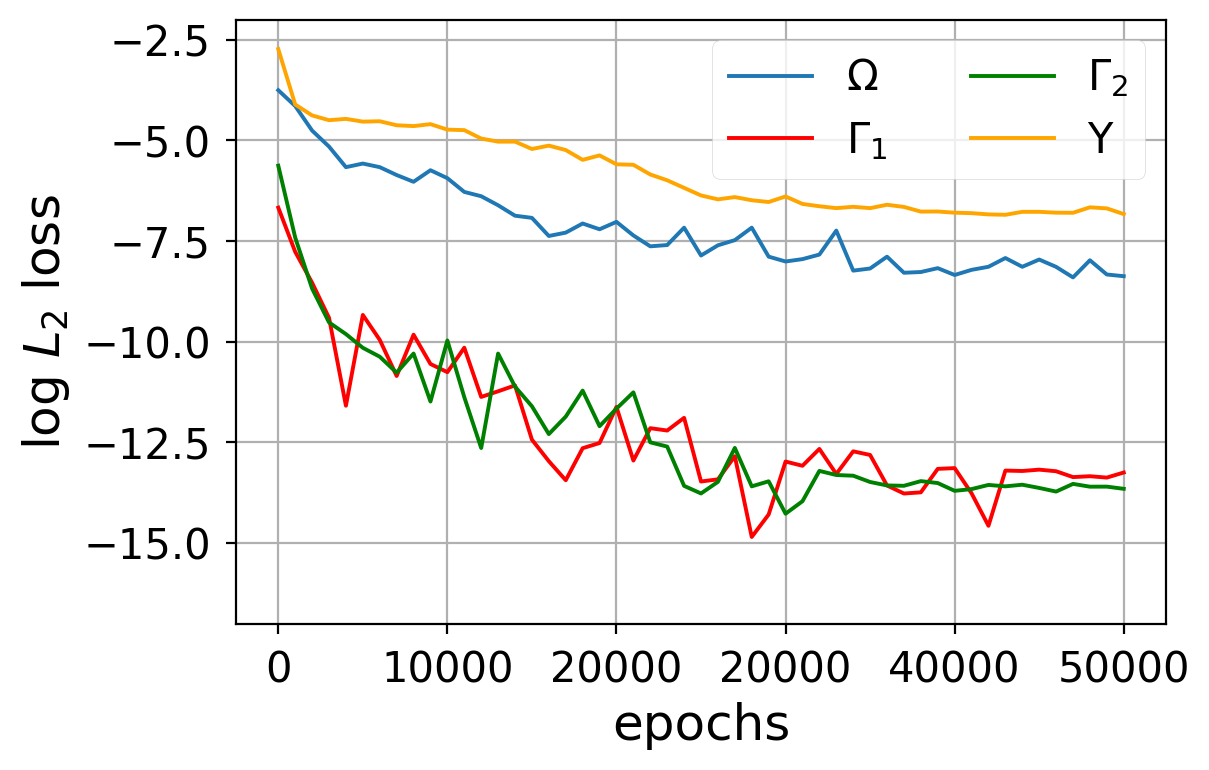}
        \includegraphics[width=0.425\linewidth]{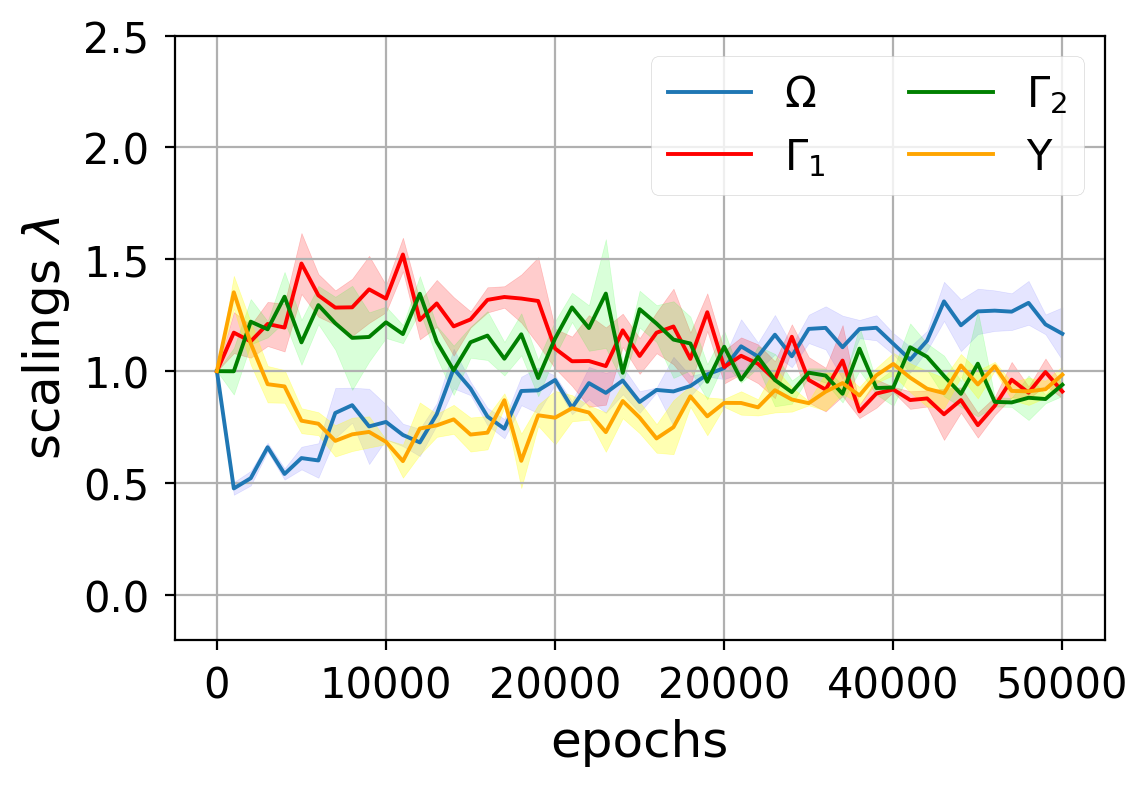}
        \caption{$\alpha = 0.9$, $\mathcal{T} = 0.1$, $\mathbb{E}[\rho] = 1$}
    \end{subfigure}
    \begin{subfigure}{\linewidth}
    \centering
        \includegraphics[width=0.45\linewidth]{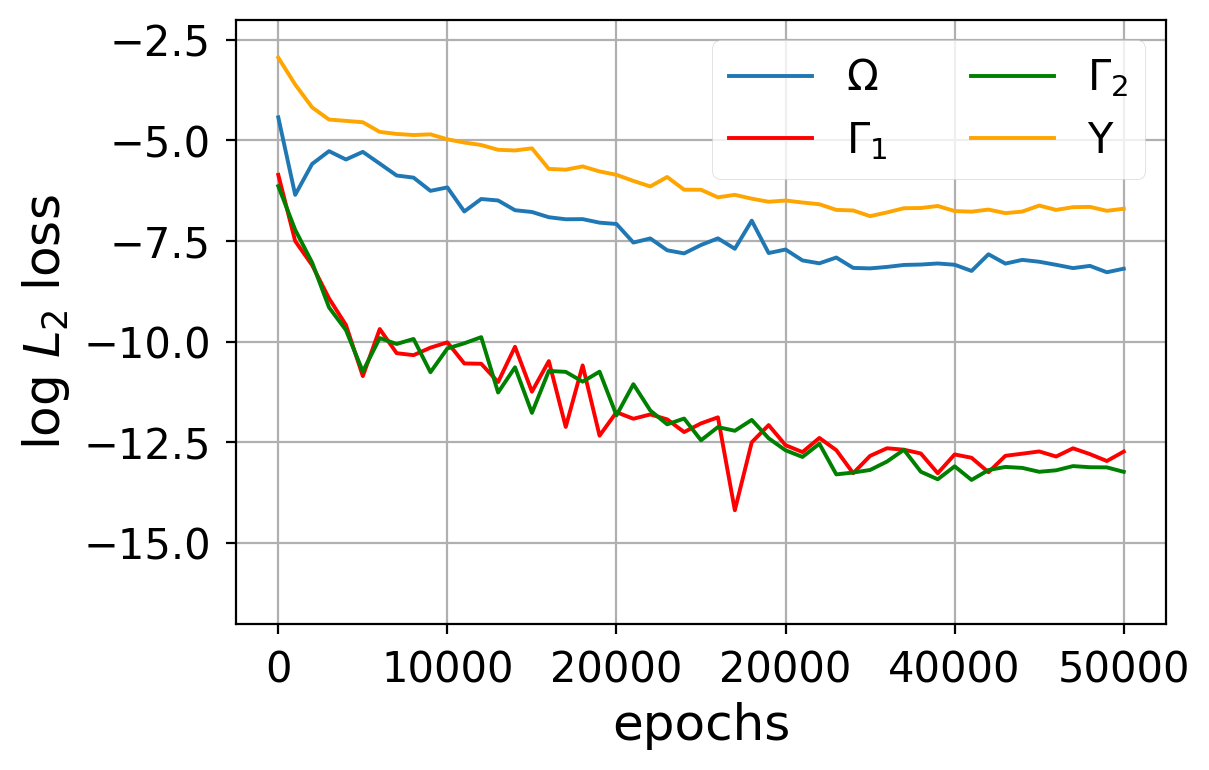}
        \includegraphics[width=0.425\linewidth]{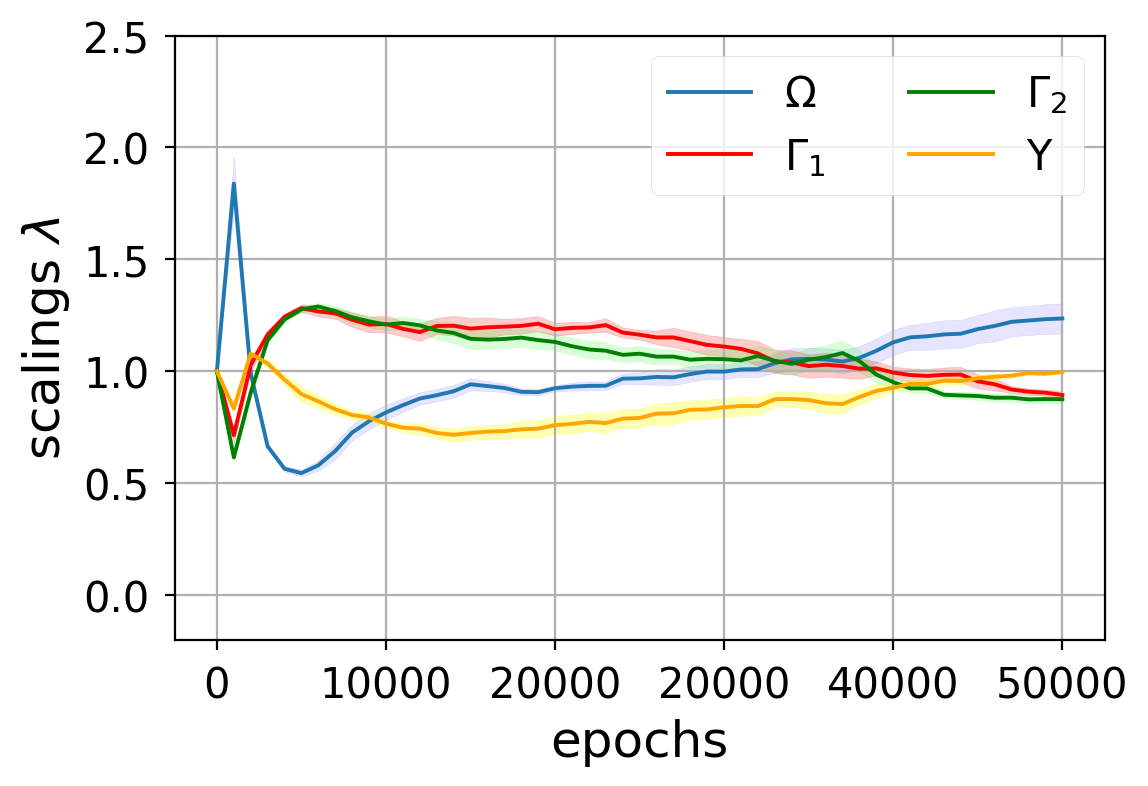}
        \caption{$\alpha = 0.999$, $\mathcal{T} = 0.1$, $\mathbb{E}[\rho] = 1$}
    \end{subfigure}
    \begin{subfigure}{\linewidth}
    \centering
        \includegraphics[width=0.45\linewidth]{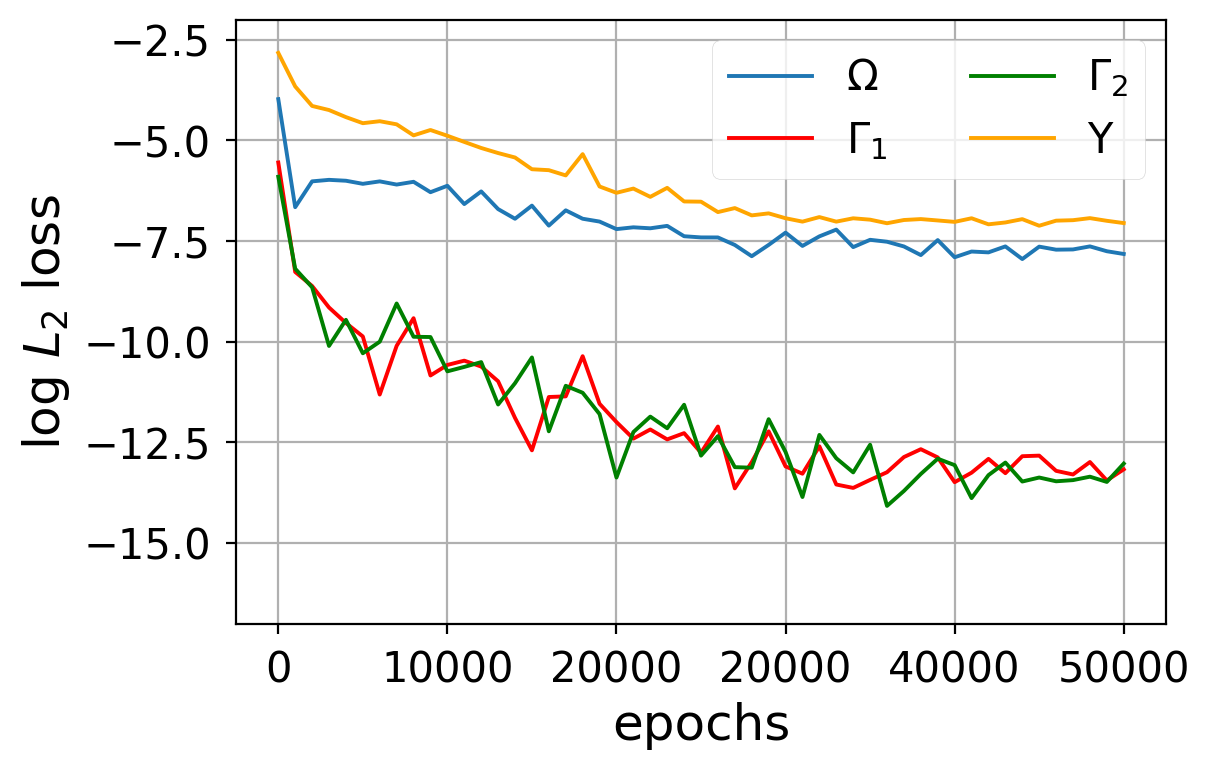}
        \includegraphics[width=0.425\linewidth]{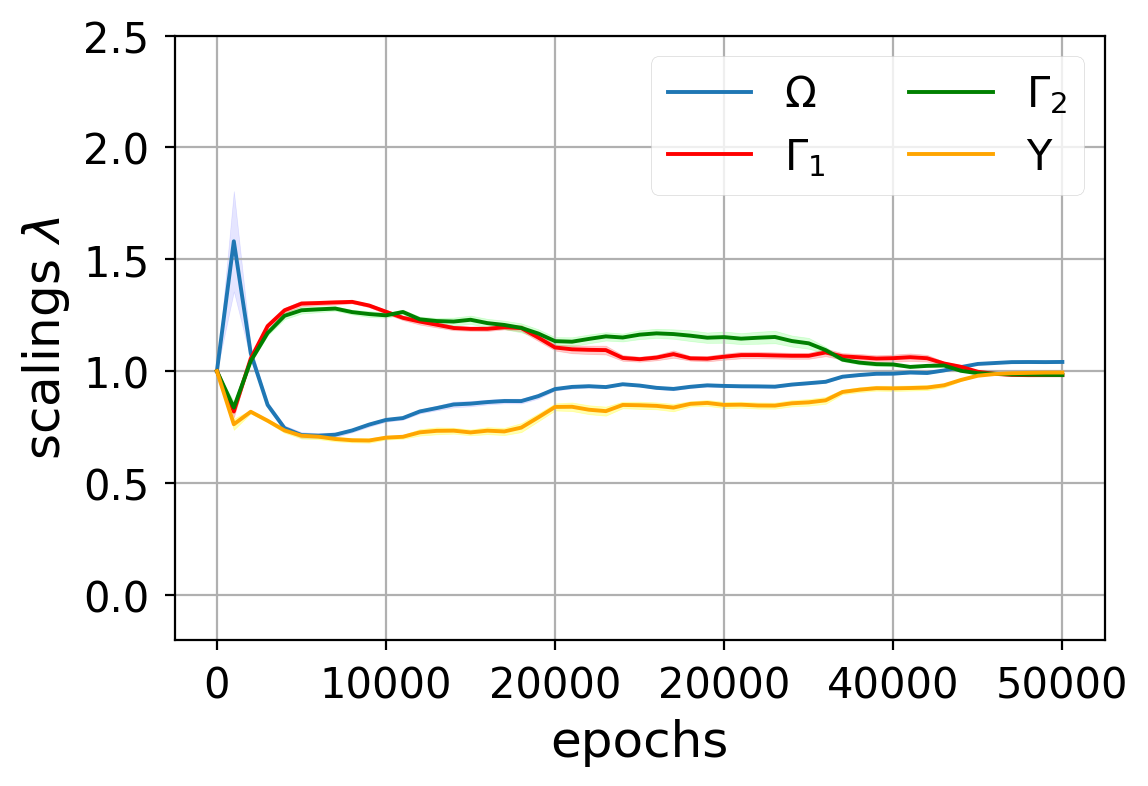}
        \caption{$\alpha = 0.999$, $\mathcal{T} = 1$, $\mathbb{E}[\rho] = 1$}
    \end{subfigure}
    \begin{subfigure}{\linewidth}
    \centering
        \includegraphics[width=0.45\linewidth]{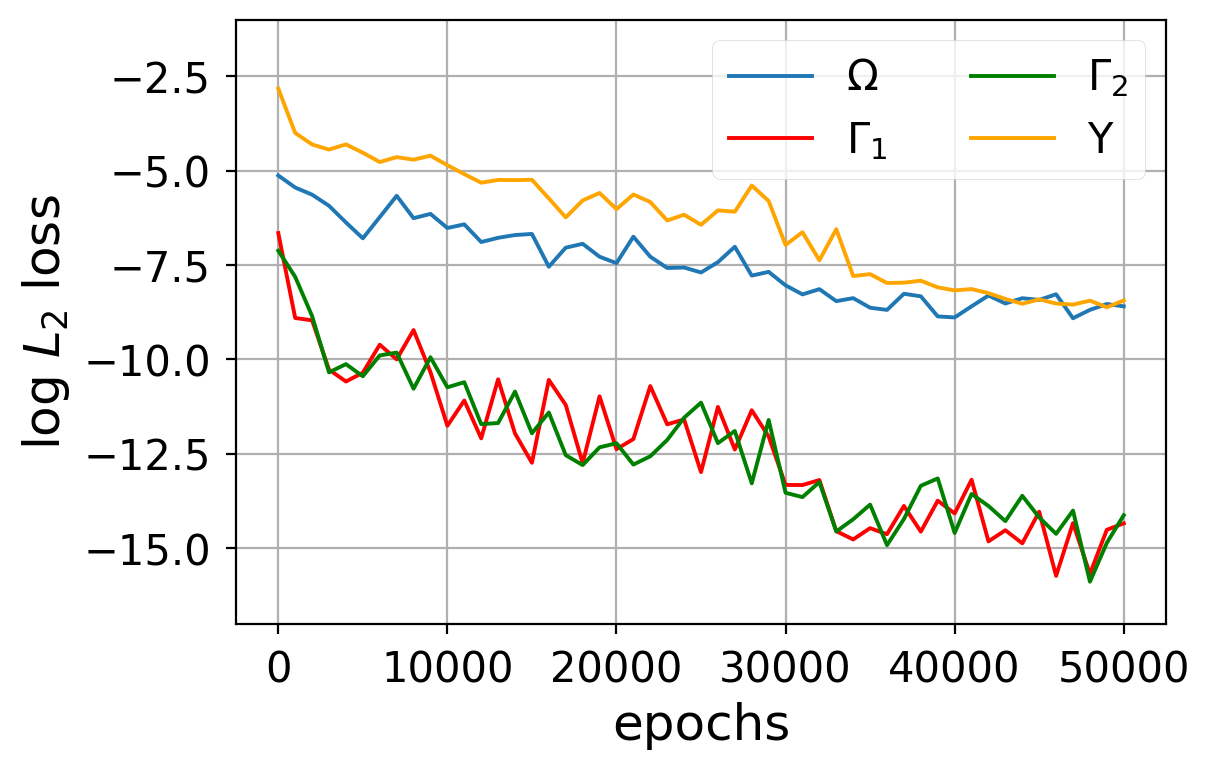}
        \includegraphics[width=0.425\linewidth]{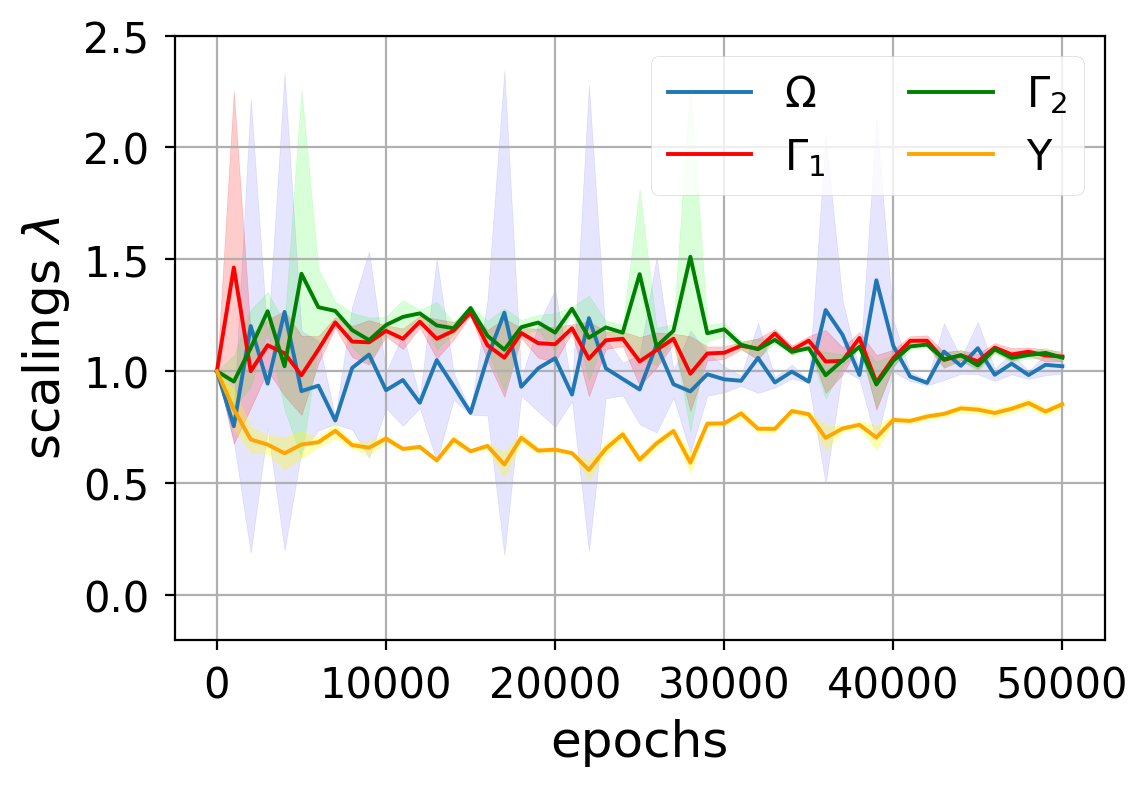}
        \caption{$\alpha = 0.999$, $\mathcal{T} = 1$, $\mathbb{E}[\rho] = 0.999$}
    \end{subfigure}
    \caption{Median of the log $L_2$ loss over multiple training runs (i/left) of Burgers' equation and the mean and variance of the corresponding scaling factors $\lambda$ (ii/right) computed with ReLoBRaLo.}
    \label{fig:loss_lambda_burgers}
\end{figure}

Fig.~\ref{fig:loss_lambda_burgers} shows the scaling factors $\lambda_i$ of our ReLoBRaLo method with varying hyperparameters for Burgers' equation. As can be expected, a larger value for $\alpha$ leads to a smoother curve because past loss statistics are dragged on longer, therefore countering the stochasticity that might arise at every optimisation step. On the other hand, the temperature $\mathcal{T}$ influences the magnitude of the scalings. Another general tendency in the plots of the scaling factors is the fact that the relatively lower loss contributions (here: BC 1 and BC 2) correspond to higher scaling values (potentially greater than 1), while the larger loss contributions (here: PDE and IC) correspond to lower scaling values (potentially less than 1). Note that, whenever "log" is used in this and all subsequent figures, we refer to the natural logarithm of that quantity.

\begin{table*}[ht]
  \centering
    \begin{tabular}{lrrrrrr}
    Burgers && Baseline & GradNorm & LR anneal. & SoftAdapt & ReLoBRaLo \\ 
    \midrule
    Forward & train $f$   & 5.5$\cdot10^{-4}$ & 6.6$\cdot10^{-4}$ & 9.9$\cdot10^{-4}$ &                2.0$\cdot10^{-4}$ & 5.6$\cdot10^{-5}$\cr
    & val $u$ & 1.2$\cdot10^{-3}$ & 2.0$\cdot10^{-3}$ & 1.6$\cdot10^{-4}$ & 8.1$\cdot10^{-4}$ & 1.4$\cdot10^{-4}$ \cr
    &std val $u$ & 5.7$\cdot10^{-4}$ & 2.1$\cdot10^{-3}$ & 2.3$\cdot10^{-4}$ & 9.5$\cdot10^{-3}$ & 6.8$\cdot10^{-4}$ \cr
    \midrule            
    Inverse & val $\mu$ & 1.9$\cdot10^{-10}$ & 6.8$\cdot10^{-5}$ & 2.5$\cdot10^{-11}$ & 1.1$\cdot10^{-7}$ & 2.2$\cdot10^{-10}$\cr
    $\nu = \frac{1}{100\pi}$ & std $\mu$ & 1.2$\cdot10^{-9}$ & 5.1$\cdot10^{-5}$ & 3.4$\cdot10^{-11}$ & 5.7$\cdot10^{-7}$ & 2.1$\cdot10^{-10}$ \cr
  \end{tabular}
  \caption{Comparison of the median $L_2$ training and validation loss on Burgers' equation against a baseline of manually chosen scalings. The reported values are the median over four independent runs with identical settings. Additionally, we report the standard deviation over the runs of the best performing model on the validation loss.}
  \label{tab:performance_balances_burgers}
\end{table*}

The relatively small variances across training runs with $\mathbb{E}[\rho] = 0$ suggest that the optimisation progress follows similar patterns, even when varying depth and width of the network. Therefore, these values can provide valuable insight into the training and help to identify possibilities of improving the model. E.g. the fact that the scaling for the governing equation has the largest value after 50,000 epochs indicates that it was the first term to stop making progress. We will see in the following sections that the opposite holds true for Helmholtz' and Kirchhoff's equations, where the boundary conditions have more difficulties making progress (cf. fig.~\ref{fig:loss_lambda_kirchhoff}). This knowledge can help taking informed decisions to improve the framework, e.g. by adapting the activation functions, the loss function or the model's architecture accordingly.

Tab.~\ref{tab:performance_balances_burgers} summarises the performances of the different balancing techniques against a baseline for the forward and inverse problem setting, where we manually chose the optimal scalings $\lambda_i$ through grid search. As can be observed, the adaptive scaling techniques perform similarly well to the baseline, with Learning Rate Annealing and ReLoBRaLo reaching a considerably lower validation error. The results show that either one of these methods greatly reduces the amount of work required for hyperparameter search, while still achieving great results with high probability.

\begin{figure}
	\centering
	\hspace{6mm}\begin{tabular}{c}
		\hline  
		\cellcolor{hellgrau} \small Convergence on inverse Burgers' problem \\
		\hline  
	\end{tabular} 
	\includegraphics[width=0.8\linewidth]{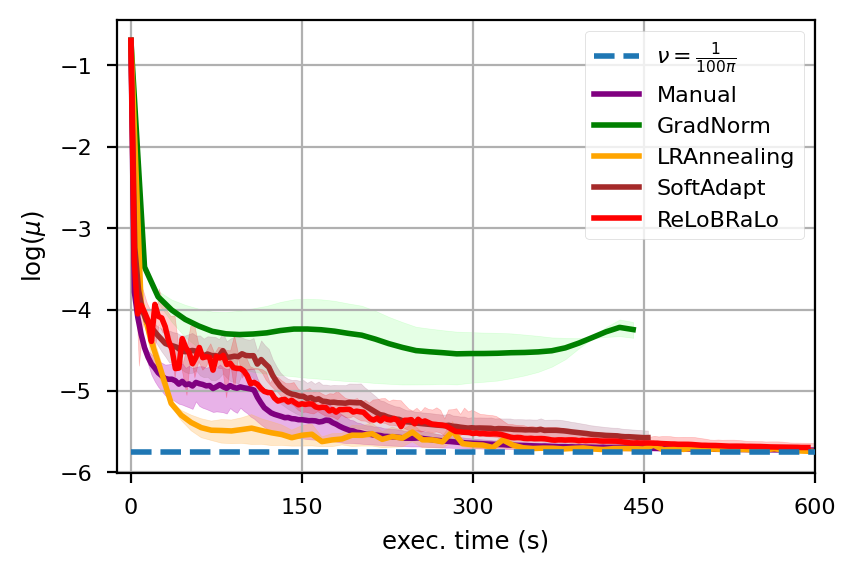} \\
     \caption{Approximation of the true PDE parameter value $\nu$ (dashed line) for the inverse problem setting of Burgers' equation. Reported values are the mean (solid line) and standard deviation (shaded area) of four independent computational runs.}
    \label{fig:mu_burgers}
\end{figure}

Besides accuracy, computational efficiency is another important metric for evaluating adaptive loss balancing methods. By designing our ReLoBRaLo method such that it requires only one backward pass, its computational overhead can be expected to be relatively small compared to GradNorm and Learning Rate Annealing, which both utilise gradient statistics and hence separate backward passes for each term. Indeed, tab.~\ref{tab:depth_width_exec_time_burgers} shows that Burgers' equation with its four terms in the loss function can be solved by ReLoBRaLo about 40\% faster than Learning Rate Annealing and 70\% faster than GradNorm and thus adds to efficiency and sustainability of PINNs training. Note that the reported values in tab.~\ref{tab:depth_width_exec_time_burgers} stem from tasks where the balancing operation was performed at every optimisation step. Both GradNorm and Learning Rate Annealing can be made more efficient by updating the scaling terms once every arbitrary number of iterations. However, this introduces a trade-off between flexibility and efficiency and therefore an additional, very sensitive hyperparameter with a high impact on the method's accuracy and efficiency. On the other hand, ReLoBRaLo adapts its scalings at every iteration and very low computational cost.

\begin{table}[h]
  \centering
    \begin{tabular}{rcccc}
     & GradNorm & LR ann. & SoftAdapt & ReLoBRaLo \\ 
    \midrule            
    $\Delta T_{co} \left[ s \right]$& +10.6 & +3.5 & +0.4 & +0.6
  \end{tabular}
  \caption{Median computational overhead $\Delta T_{co}$ (in s) per 1'000 optimisation steps compared to using no balancing scheme (3.7s) on Burgers' equation.}
  \label{tab:depth_width_exec_time_burgers}
\end{table}

It is noteworthy that, while the forward problem induced a loss function consisting of four terms, the inverse problem requires only two terms (Eq.~\ref{eq:burgers_inverse_loss}). Hence, selecting the scalings manually is significantly less time-consuming than it is for the forward problem. Consequently, tab.~\ref{tab:performance_balances_burgers} shows that the baseline was harder to outperform, with Learning Rate Annealing (LR Annealing) and ReLoBRaLo being the only methods yielding better results. It is worth noting however that Learning Rate Annealing approximates the true value of $\nu$ significantly faster than ReLoBRaLo and is therefore the optimal choice for this particular problem setting, cf. fig.~\ref{fig:mu_burgers}. Further conclusions and comparisons across different loss balancing methods are made in sec.~\ref{sec:Ablation}.

\begin{figure*}[h]
	\centering
	\begin{tabular}{c c c}
		\hline  
		\hspace{5mm} \cellcolor{hellgrau} (a) Analytical Result $\left[ m \right]$  &\hspace{7mm}  \cellcolor{hellgrau} \small (b) PINN-Result $\left[ m \right]$ \hspace{10mm} & \cellcolor{hellgrau} \small (c) Squared Error $\left[ m^2 \right]$ \hspace{8mm} \\ 
		\hline  
		\vspace{-2mm}
	\end{tabular} \\
	\includegraphics[width=0.25\linewidth]{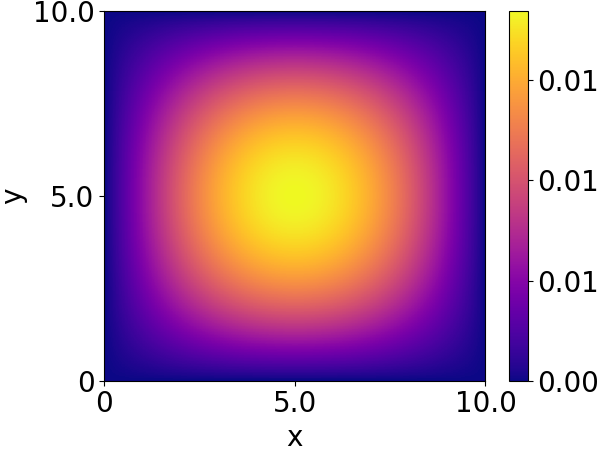} \includegraphics[width=0.25\linewidth]{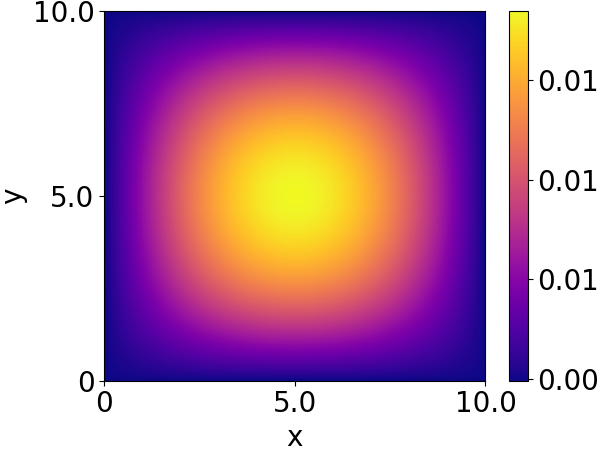} \includegraphics[width=0.278\linewidth]{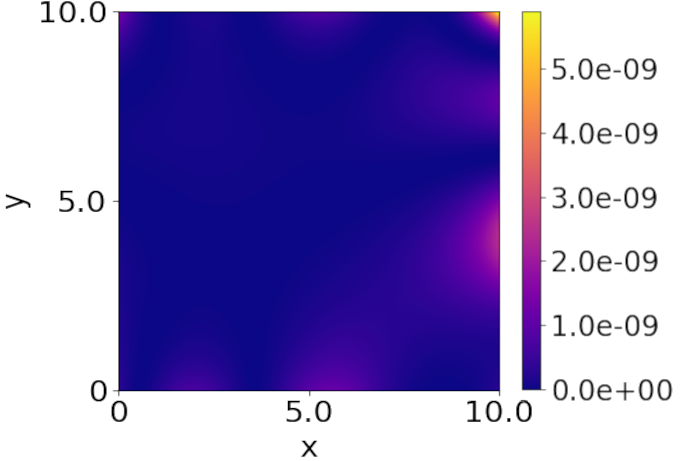} \\
    \caption{Kirchhoff plate bending problem: (a) analytical reference solution $\left[ m \right]$, (b) PINNs results $\left[ m \right]$ predicted with a fully-connected network consisting of three layers and 128 nodes each, and (c) squared error $\left[ m^2 \right]$.}
    \label{fig:pred_kirchhoff}
\end{figure*}

\subsection{Kirchhoff Plate Bending Equation}
The Kirchhoff–Love theory of plates arose from civil and mechanical engineering and consists of a two-dimensional mathematical model used to determine stresses and deformations in thin plates subjected to forces and moments \cite{bathe1996finite}. The Kirchhoff plate bending problem assumes that a mid-surface plane can be used to represent a three-dimensional plate in two-dimensional form and together with a linear elastic material a fourth-order PDE can be derived to describe its mechanical behaviour:
\begin{equation}
    \label{eq:kirchhoff}
    \begin{aligned}
        &\nabla^4 u(x, y) - \frac{p(x, y)}{D} = 0, \quad (x, y) \in \mathbb{R}_{>0}^2\\
        &D = \frac{E h^3}{12(1-\nu^2)}
    \end{aligned}
\end{equation}
where $p(x, y)$ is the load acting on the plate at coordinates $(x, y)$; $D$ is the plate's flexural stiffness computed with Young's modulus $E$, the plate's thickness $h$ and Poisson's ratio $\nu$. The Kirchhoff plate bending problem poses several severe problems to FEM solutions \cite{bathe1996finite}, yet analytical solutions can be inferred e.g. using Fourier series for special cases such as an applied sinusoidal load:
\begin{equation}
    \label{eq:kirchhoff_analytical}
    \begin{aligned}
        &p(x, y) = p_0\sin\left(\frac{x\pi}{a}\right)\sin\left(\frac{y\pi}{b}\right)\\
        &u(x, y) = \frac{p_0}{\pi^4 D (\frac{1}{a^2} + \frac{1}{b^2})^2}\sin\left(\frac{x\pi}{a}\right)\sin\left(\frac{y\pi}{b}\right)
    \end{aligned}
\end{equation}
In this paper we consider a concrete plate of width $a =$ \SI{10}{\meter}, length $b =$ \SI{10}{\meter}, base load $p_0 =$ \SI{0.015}{\mega\newton\per\meter\squared}, Young's modulus $E =$ \SI{30,000}{\mega\newton\per\meter\squared}, plate height $h =$ \SI{0.2}{\meter} and Poisson's ratio of $\nu = 0.2$, as well as simply supported edge boundary conditions as it arises in typical civil engineering structures such as slabs \cite{concrete_structures}. We hence consider the following boundary conditions (BC): 
\begin{equation}
    \label{eq:kirchhoff_boundary_conditions}
    \begin{aligned}
        &u(0, y) = u(a, y) = u(x, 0) = u(x, b) = 0\\
        &m_{x}(0, y) = m_{x}(a, y) = m_{y}(x, 0) = m_{y}(x, b) = 0
    \end{aligned}
\end{equation}
where $m_x$ and $m_y$ are bending moments computed as follows:
\begin{equation}
    \label{eq:bending_moments}
    \begin{aligned}
        &m_{x}(x, y) = -D \left(\partial^2_x u + \nu \partial^2_y u\right)\\
        &m_{y}(x, y) = -D \left(\nu \partial^2_x u + \partial^2_y u\right)\\
        &m_{xy}(x, y) = -D(1-\nu) \partial^2_{xy} u \\
    \end{aligned}
\end{equation}

In total, we obtain 8 boundary conditions and therefore 9 terms in the PINNs loss function, making this a challenging task for balancing the contributions of the various objectives:
\begin{equation}
    \label{eq:kirchhoff_loss}
    \begin{aligned}
        \mathcal{L}^{(t)} &= \frac{\lambda_0}{\vert \hat{\Omega} \vert}\sum_{x, y \in \hat{\Omega}}{ \left\lVert \nabla^4 U(x, y; \mathbf{\theta}) - \frac{p}{D} \right\rVert_2^2 } \\
        &+ \sum_{i = 1}^4 \frac{\lambda_i}{\vert \hat{\Gamma}_i \vert}\sum_{x,y \in \hat{\Gamma}_i}\left\lVert U(x, y; \mathbf{\theta}) \right \rVert_2^2\\  
        &+ \sum_{i = 5}^6 \frac{\lambda_i}{\vert \hat{\Gamma}_i \vert}\sum_{x,y \in \hat{\Gamma}_i}{ \left\lVert m_{ux}(x, y) \right\rVert_2^2 } \\
        &+ \sum_{i = 7}^8 \frac{\lambda_i}{\vert \hat{\Gamma}_i \vert}\sum_{x,y \in \hat{\Gamma}_i}{ \left\lVert m_{uy}(x, y) \right\rVert_2^2 } 
    \end{aligned}
\end{equation}
\begin{figure} [ht]
	\centering
	\begin{tabular}{c c}
		\hline  
	\hspace{5mm} \cellcolor{hellgrau} \small (i) $L_2$ Convergence  & \hspace{6mm} \cellcolor{hellgrau} \small (ii) ReLoBRaLo Scaling \hspace{1mm}\\ 
		\hline  
	\end{tabular} 
	\includegraphics[width=0.49\linewidth]{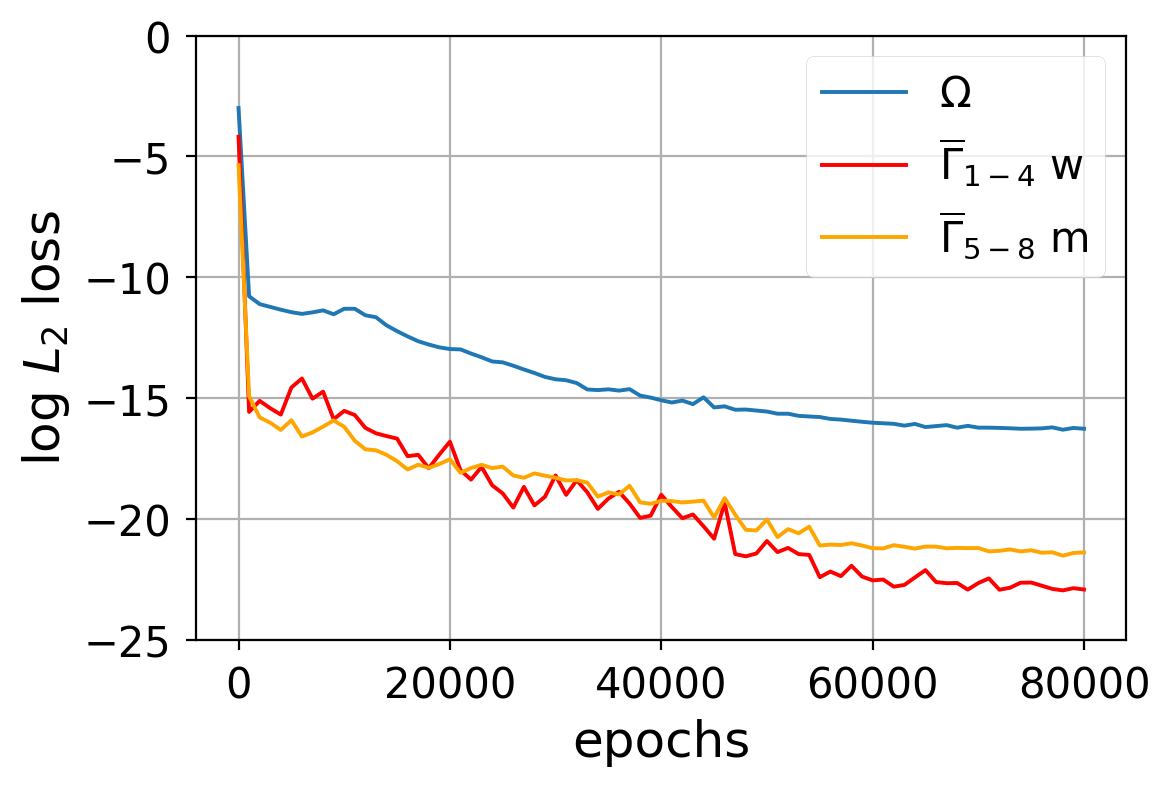} \includegraphics[width=0.482\linewidth]{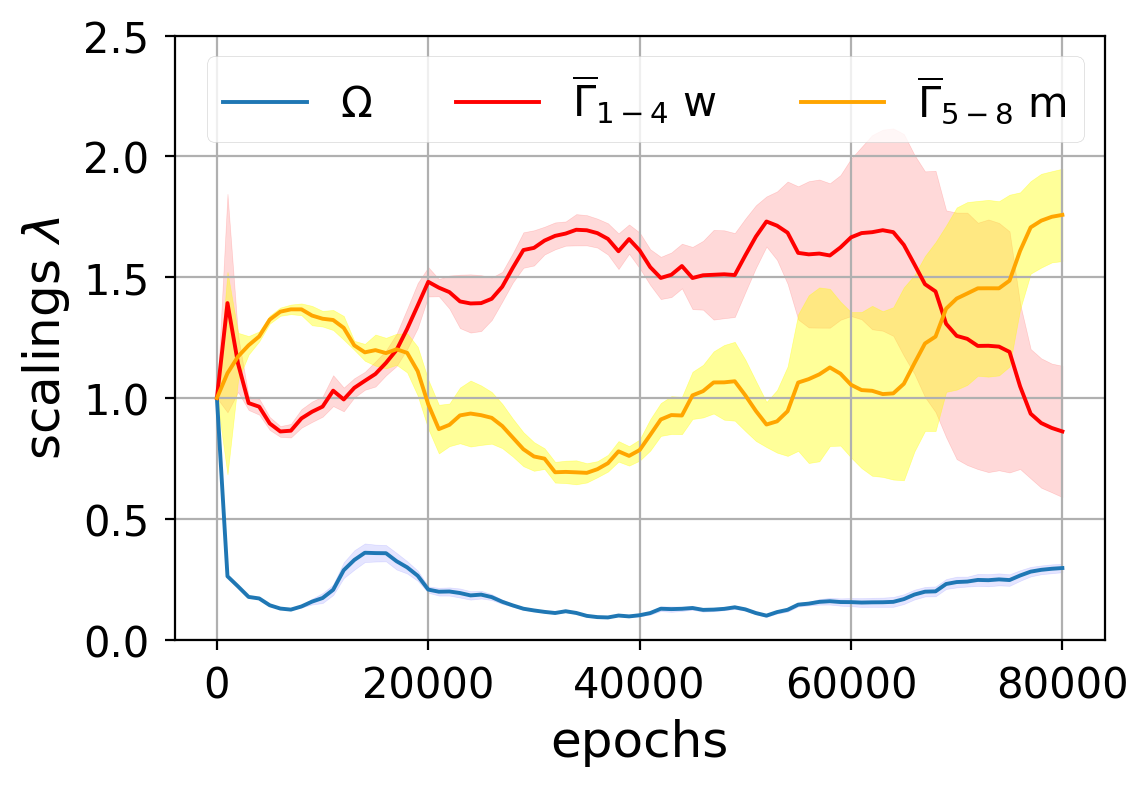}
    \caption{Median of the log $L_2$ loss over multiple training runs (a) and the mean and variance of the corresponding scaling factors $\lambda$ (b) computed with ReLoBRaLo on Kirchhoff's equation with $\alpha = 0.999$, $\mathcal{T} = 0.1$, $\mathbb{E}[\rho] = 1$. For the sake of readability, the boundary conditions $1-4$ and $5-8$ were aggregated by taking the mean value.}
    \label{fig:loss_lambda_kirchhoff}
\end{figure}
After the successful convergence of the PINNs training, we obtain the results displayed in fig.~\ref{fig:pred_kirchhoff}(b) and compare it to the analytically available solution displayed in fig.~\ref{fig:pred_kirchhoff}(a). A plot of the squared difference in $u$ as given by the analytical and PINNs results is shown in fig.~\ref{fig:pred_kirchhoff}(c) and delivers a negligible maximum error. The final algorithm settings are reported in tab.~\ref{tab:bayes_opt_parameters_final}. 

Fig.~\ref{fig:loss_lambda_kirchhoff} shows an example of ReLoBRaLo's training progress on Kirchhoff's equation. In this particular example, one can notice the larger variance of scaling values towards the end of training. Also, the scalings did not converge towards the value 1, thus suggesting that the training finished without all terms having stopped making progress, i.e. the scalings for the boundary conditions on the moments (yellow) were increasing at the end of training, while the boundary conditions on the displacements (red) were decreasing. This gives a strong indication as to where the model's limitations lie. In this case, additional attention should be paid to the moments, e.g. by selecting an activation function which is better behaved in the second derivative than $\tanh$. Note how ReLoBRaLo in combination with early stopping weakly imposes Pareto optimal updates, as it gradually increases the weights of underperforming terms and eventually leads to a stop of the training process due to a lack of global progress. This is an important property in the context of PINNs, because optimising only a subset of terms in the loss can lead to unsatisfactory solutions from a physical perspective.
\begin{table*}
  \centering
    \begin{tabular}{lrrrrrr}
    Kirchhoff & & Baseline & GradNorm & LR anneal. & SoftAdapt & ReLoBRaLo \\ 
    \midrule            
    Forward   &  train $f$   & 1.2$\cdot10^{-7}$ & 5.3$\cdot10^{-7}$ & 9.1$\cdot10^{-9}$ &                1.8$\cdot10^{-8}$ & 6.0$\cdot10^{-9}$ \cr
                &  val $u$ & 1.3$\cdot10^{-8}$ & 1.7$\cdot10^{-8}$ & 2.7$\cdot10^{-9}$ & 2.5$\cdot10^{-9}$ & 4.0$\cdot10^{-10}$ \cr
                &  std val $u$ & 3.9$\cdot10^{-8}$ & 2.2$\cdot10^{-7}$ & 1.0$\cdot10^{-6}$ & 1.9$\cdot10^{-9}$ & 7.7$\cdot10^{-10}$ \cr
    \cmidrule{1-7}
    Inverse   &  val $\mu$   & 2.1 & 3.6 & 6.0 & 9.5 & 3.2$\cdot10^{-2}$ \cr
    $D = 20.8\overline{3}$  &  std $\mu$ & 1.6 & 4.7 & 0.8 & 4.9 & 2.9$\cdot10^{-2}$
  \end{tabular}
  \caption{Comparison of the median $L_2$ training and validation loss on Kirchhoff's equation against a baseline of manually chosen scalings. The reported values are the median over four independent runs with identical settings. Additionally, we report the standard deviation over the runs of the best performing model on the validation loss.}
  \label{tab:performance_balances_kirchhoff}
\end{table*}

\begin{figure}
	\centering
	\hspace{7mm}\begin{tabular}{c}
		\hline  
		\cellcolor{hellgrau} \small Convergence on inverse Kirchhoff problem \\
		\hline  
	\end{tabular} 
	\includegraphics[width=0.8\linewidth]{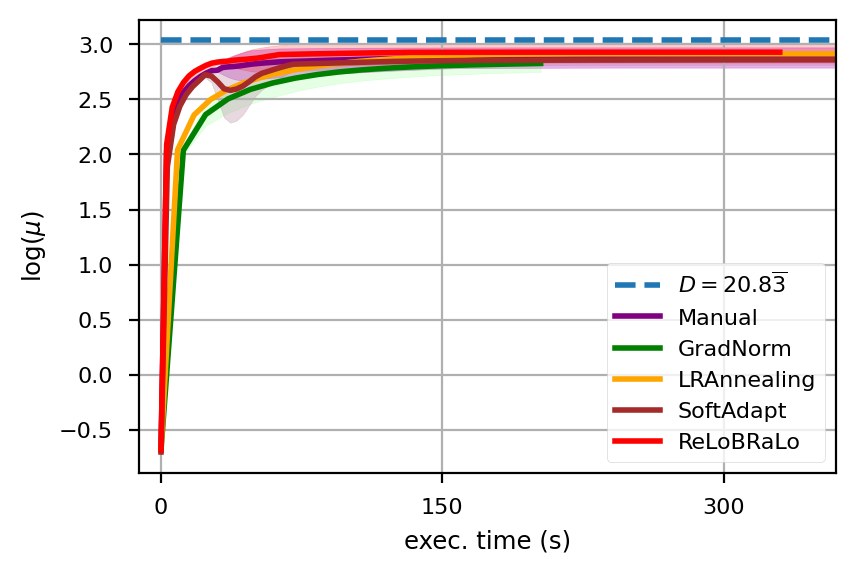}
    \caption{Approximation of the true PDE parameter value $D$ (dashed line) for the inverse problem setting of Kirchhoff plate bending. Reported values are the mean (solid line) and standard deviation (shaded area) of four independent runs.}
    \label{fig:mu_kirchhoff}
\end{figure}

Concerning performance, ReLoBRaLo outperforms the baseline and other algorithms by almost an order of magnitude in accuracy, while also yielding a very small standard deviation and hence being very consistent across training runs (cf. tab.~\ref{tab:performance_balances_kirchhoff}). The results show its effectiveness, even on Kirchhoff's challenging problem with a total of 9 terms (cf. Eq.~\ref{eq:kirchhoff_loss}). Furthermore, the execution times in tab~\ref{tab:exec_time_kirchhoff} underline the efficiency benefit (up to sixfold speedup) of balancing the loss without gradient statistics, as separate backwards passes for each term become increasingly computationally expensive as the number of terms in the loss function grows. Further conclusions and comparisons across different loss balancing methods are made in sec.~\ref{sec:Ablation}.

\begin{table}
  \centering
    \begin{tabular}{rcccc}
     & GradNorm & LR ann. & SoftAdapt & ReLoBRaLo \\ 
    \midrule            
    $\Delta T_{co} \left[ s \right]$ & 128.6 & 139.7 & 20.2 & 22.5
  \end{tabular}
  \caption{Median computational overhead $\Delta T_{co}$ (in s) per 1'000 optimisation steps compared to using no balancing scheme (17.3s) on Kirchhoff's equation.}
  \label{tab:exec_time_kirchhoff}
\end{table}

For the inverse Kirchhoff problem setting, we select the PDE parameter $\mu := D$ (i.e. flexural stiffness) to be learned for given data, which we obtained by sampling from the analytically known solution. More specifically, we initialised $\mu = 0.5$ and tasked the network with approximating $D = 20.8\overline{3}$. Given the large disparity between the initialisation and the target, we empirically found the use of two separate optimisers beneficial in this case, where one optimiser is used for updating the network's parameters $\theta$ and a different one for updating the PDE parameter $\mu$. Differently from Burgers' equation, ReLoBRaLo also sets a new benchmark in Kirchhoff's inverse problem, both in accuracy as well as convergence speed, cf. fig.~\ref{fig:mu_kirchhoff} and tab.~\ref{tab:exec_time_kirchhoff}.

\subsection{Helmholtz equation}\label{sec:helmholtz}
The Helmholtz equation represents a time-independent form of the wave equation and arises in many physical and engineering problems such as acoustics and electromagnetism \citep{pde_sommerfeld}. The equation has the form:
\begin{equation}
    \label{eq:helmholtz}
    \Delta u(x, y) + k^2u(x, y) = f(x, y), \quad x, y \in [-1, 1]^2
\end{equation}
where $k$ is the wave number. This represents a common problem to benchmark PINNs and possesses an analytical solution in combination with Dirichlet boundaries:
\begin{equation}
    \label{eq:helmholtz_analytical}
    \begin{aligned}
        &f(x, y) = (-\pi^2 - (4\pi)^2 + k^2) sin(\pi x) sin(4 \pi y) \\
        &u(x, y) = sin(\pi x) sin(4 \pi y)\\
        &u(-1, y) = u(1, y) = u(x, -1) = u(x, 1) = 0
    \end{aligned}
\end{equation}
\begin{figure*}[ht]
	\centering
	\begin{tabular}{c c c}
		\hline  
		\cellcolor{hellgrau} \small (a) Analytical Result  & \cellcolor{hellgrau} \small (b) PINN-Result & \cellcolor{hellgrau} \small (c) Squared Error \\ 
		\hline  
		\vspace{.5mm}
		\includegraphics[width=0.255\linewidth]{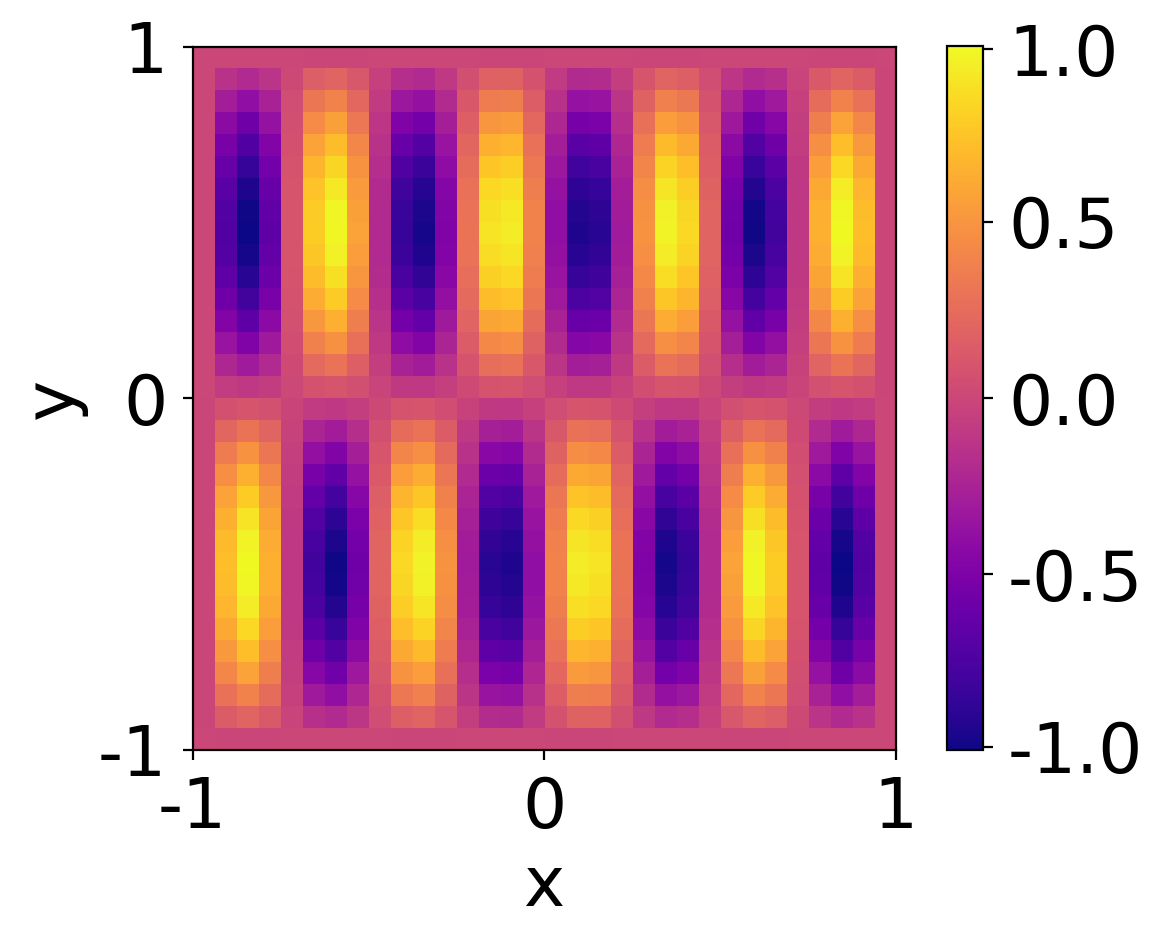} & \includegraphics[width=0.255\linewidth]{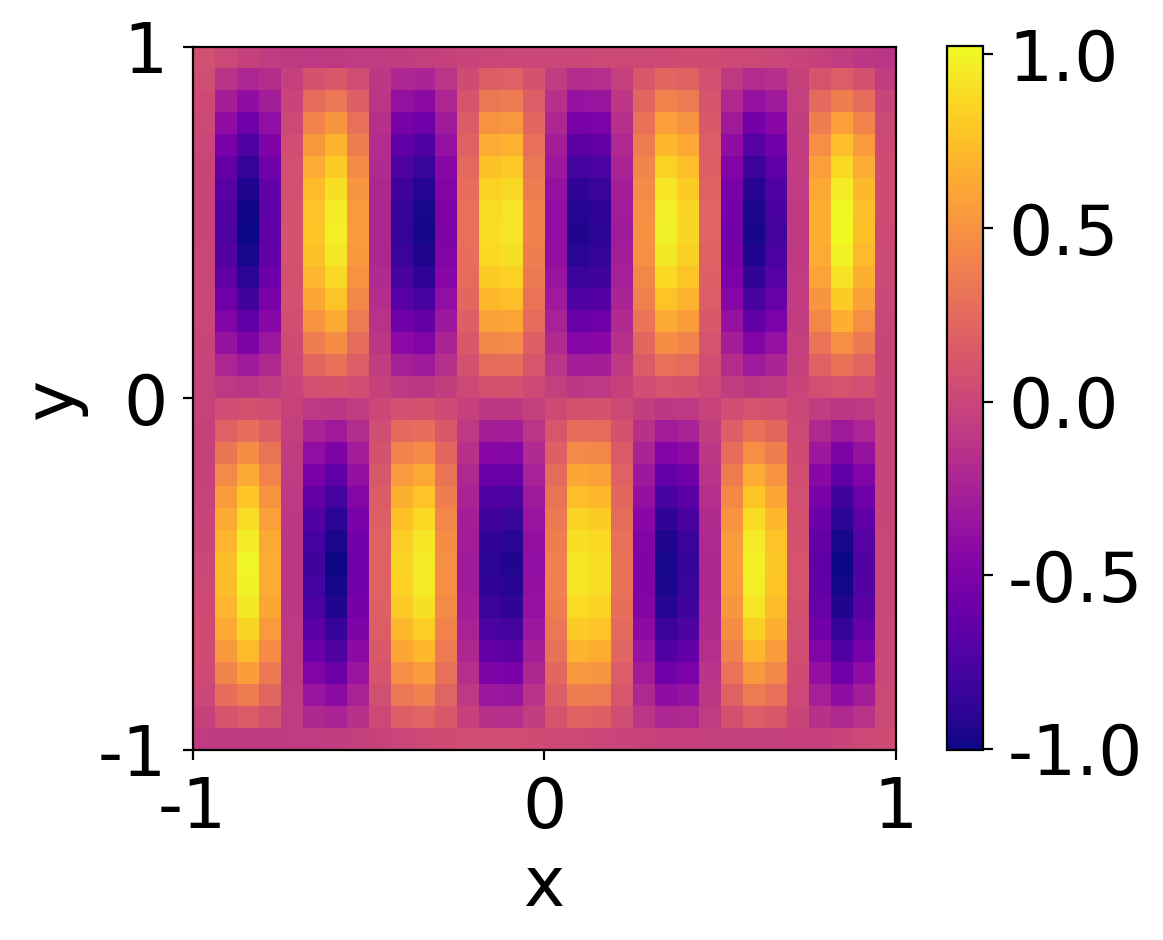}  & \includegraphics[width=0.285\linewidth]{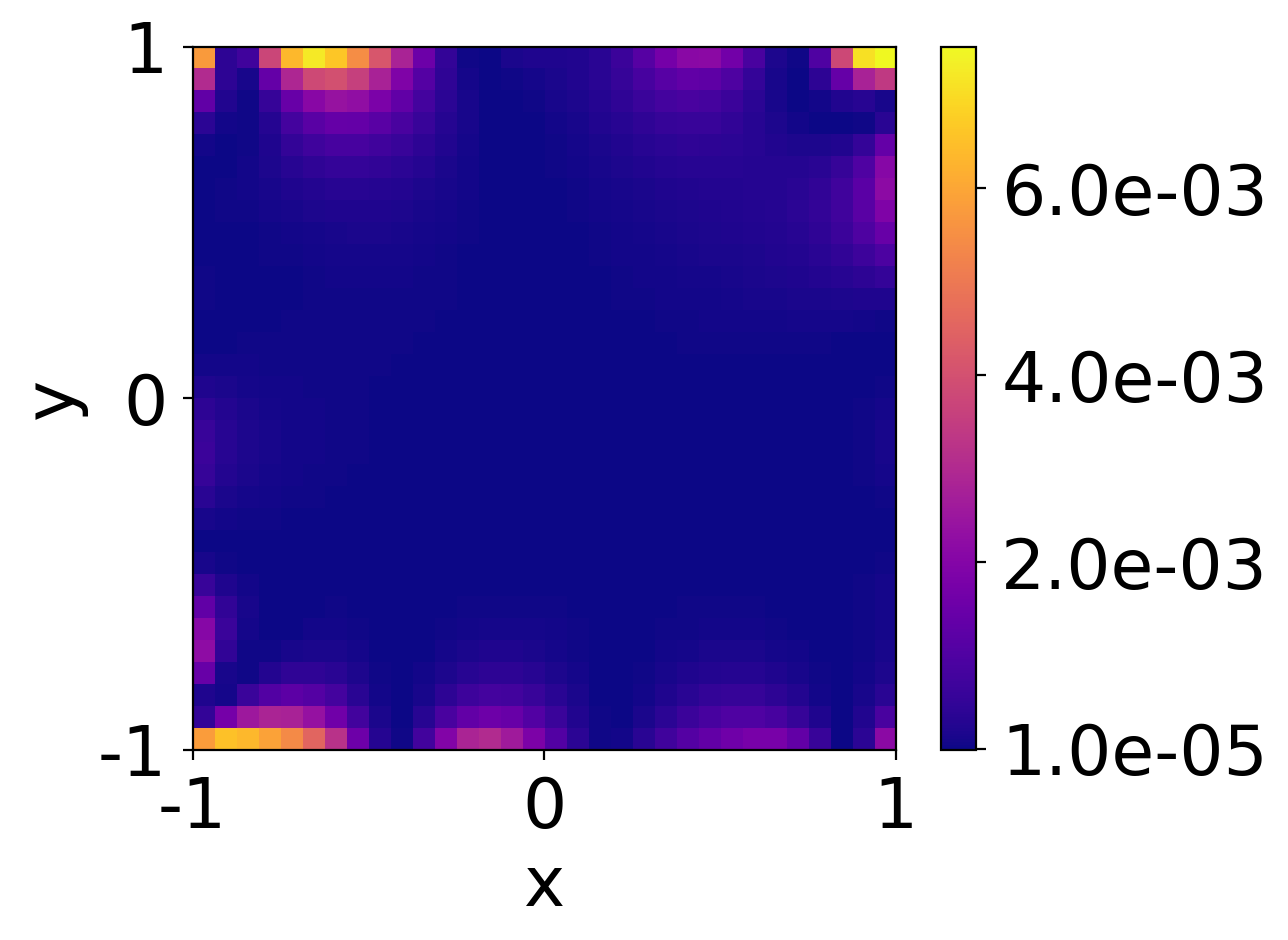} \\
	\end{tabular} 
    \caption{Helmholtz's problem: (a) analytical reference solution, (b) PINNs results predicted with a fully-connected network consisting of two layers and 128 nodes each, and (c) squared error.}
    \label{fig:pred_helmholtz}
\end{figure*}

Both, the $x_1$ and $x_2$ input variables, are bounded below by -1 and bounded above by 1. Therefore, the boundary conditions add four terms to the loss function of the \textit{forward problem}, resulting in a 5-term total physics-informed loss:
\begin{equation}
    \label{eq:helmholtz_loss}
    \begin{aligned}
        \mathcal{L}^{(t)} &= \frac{\lambda_0}{\vert \hat{\Omega} \vert}\sum_{x,y \in \hat{\Omega}}{ \left\lVert \Delta U(x,y; \mathbf{\theta}) + k^2U(x,y; \mathbf{\theta}) - f(x,y) \right\rVert_2^2 }\\
        &+ \sum_{i = 1}^4 \frac{\lambda_i}{\vert \hat{\Gamma}_i \vert}\sum_{x,y \in \hat{\Gamma}_i}{ \left\lVert U(x,y; \mathbf{\theta}) \right\rVert_2^2 } 
    \end{aligned}
\end{equation}
where $U(x, y; \mathbf{\theta})$ is the parameterisation of the latent function $\hat{\mathbf{u}}(x,y)$ using a neural network with parameters $\mathbf{\theta}$.

After a successful convergence of the PINNs training, we obtain the results displayed in fig.~\ref{fig:pred_helmholtz}(b) and compare it to the analytically available solution displayed in fig.~\ref{fig:pred_helmholtz}(a). A plot of the squared difference in $u$ as given by the analytical and PINNs results is shown in fig.~\ref{fig:pred_helmholtz}(c) and delivers a negligible max error. The final algorithm settings are reported in tab.~\ref{tab:bayes_opt_parameters_final}. 

\begin{figure}
	\centering
	\begin{tabular}{c c}
		\hline  
		\hspace{5mm}\cellcolor{hellgrau} \small (a) GradNorm Result  & \hspace{2mm} \cellcolor{hellgrau} \small (b) LR Annealing Result \\
		\hline 
	\end{tabular}
	\includegraphics[width=0.49\linewidth]{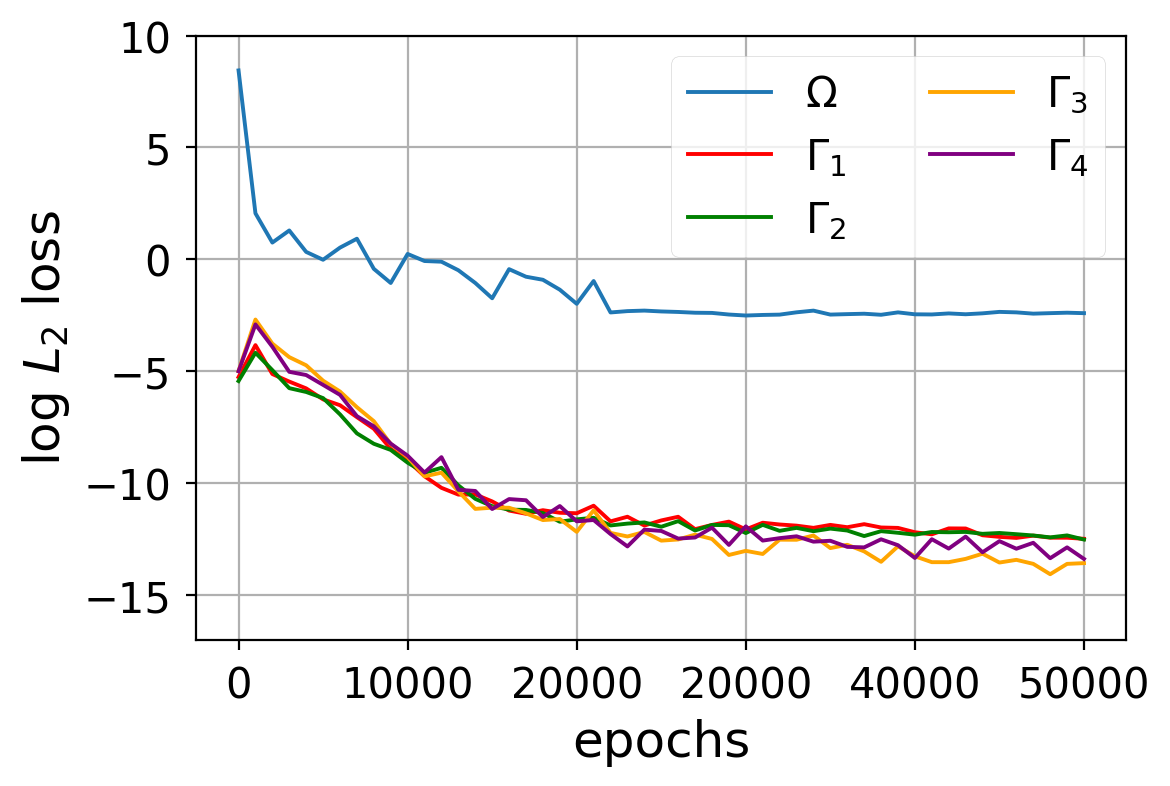} \includegraphics[width=0.49\linewidth]{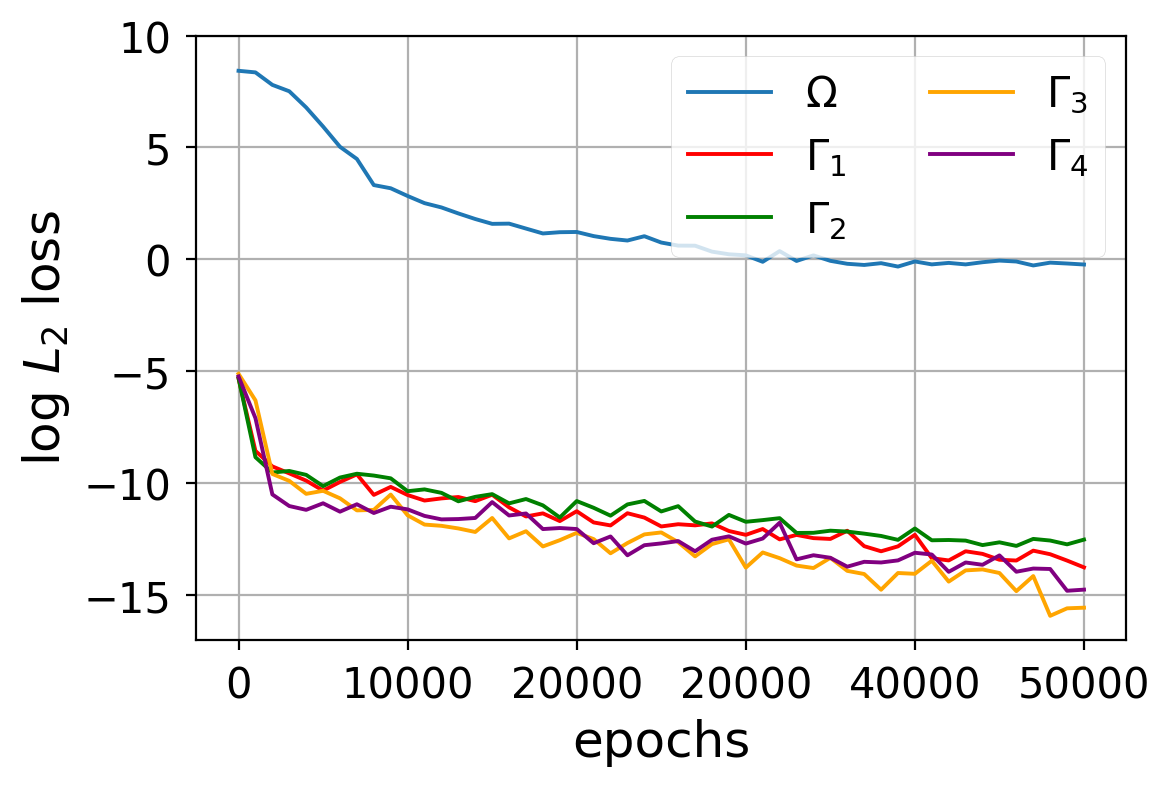}
	\begin{tabular}{c c}
		\hline  
		\hspace{5mm}\cellcolor{hellgrau} \small (c) ReLoBRaLo Result  & \hspace{2mm} \cellcolor{hellgrau} \small (d) ReLoBRaLo Scaling\\
		\hline  
	\end{tabular}
	\includegraphics[width=0.49\linewidth]{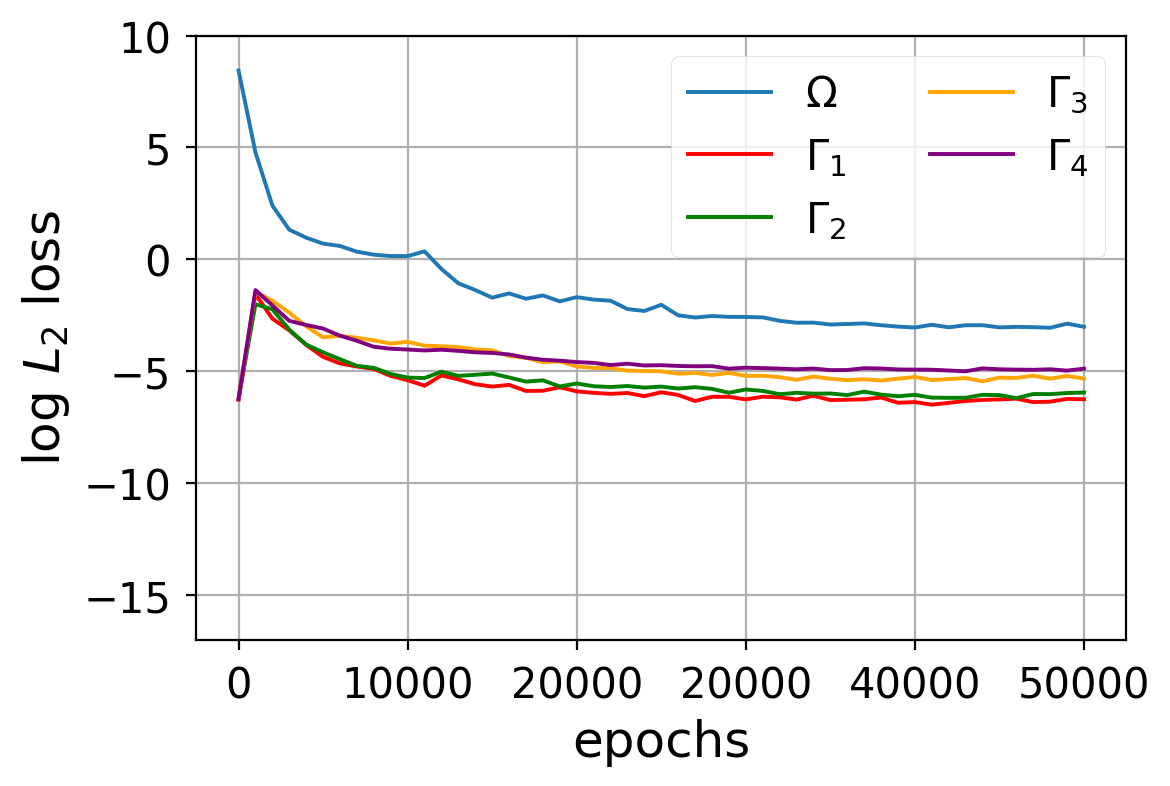} \includegraphics[width=0.485\linewidth]{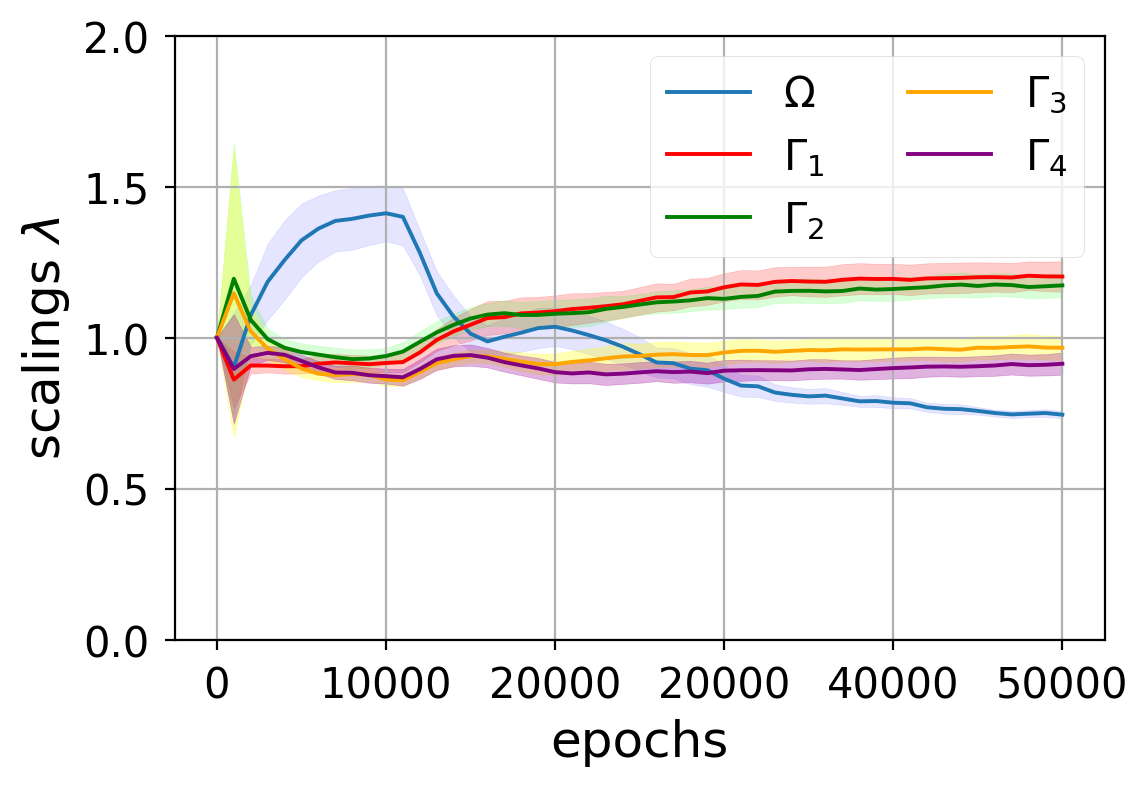}
    \caption{Median of the log $L_2$ loss over multiple training runs on the Helmholtz's equation using (a) GradNorm, (b) Learning Rate Annealing and (c) ReLoBRaLo, as well as the mean and variance of scalings calculated by ReLoBRaLo (d).}
    \label{fig:loss_helmholtz}
\end{figure}

The Helmholtz equation reveals a limitation of our basic loss balancing approach and motivates the introduction of the random lookback. GradNorm and Learning Rate Annealing both achieve impressive results and substantially outperform the baseline as well as ReLoBRaLo with $\mathbb{E}[\rho] = 1$ in terms of $L2$ accuracy for the BC terms, cf. fig.~\ref{fig:loss_helmholtz}. This is likely due to the considerable initial difference in magnitudes between the governing equation and the boundary conditions. Furthermore, the high values of $\alpha$, necessary for "remembering" the deteriorations longer, induce a latency between the increase of a term's loss until the scaling $\lambda$ reacts accordingly (cf. fig.~\ref{fig:loss_helmholtz}(d)). On the other hand, GradNorm and Learning Rate Annealing do not succeed in decreasing the $L2$ error as much as ReLoBRaLo for the governing equation term. Fig.~\ref{fig:loss_helmholtz} shows that both GradNorm and Learning Rate Annealing focus on improving the boundary conditions right from the beginning of training, whereas ReLoBRaLo with $\mathbb{E}[\rho] = 1$ counters the initial deterioration, but eventually "forgets" and instead focuses on the more dominant governing equation. This is also reflected in the discrepancy between the training and validation loss: GradNorm and Learning Rate Annealing have a higher training loss than ReLoBRaLo, but still exceed at approximating the underlying function (cf. tab.~\ref{tab:performance_balances_helmholtz}). This triggered further investigation on the saudade and temperature parameters as described in the remainder of the next section.

\begin{table*}[h]
  \centering
    \begin{tabular}{lrrrrrr}
    Helmholtz      &  & Baseline & GradNorm & LR anneal. & SoftAdapt & ReLoBRaLo \\ 
    \midrule            
    Forward   &  train $f$      & 1.4$\cdot10^{-2}$ & 7.1$\cdot10^{-2}$ & 2.7$\cdot10^{-1}$ &              9.5$\cdot10^{-3}$ & 4.7$\cdot10^{-3}$ \cr
                &  val $u$ & 7.1$\cdot10^{-2}$ & 5.6$\cdot10^{-6}$ & 1.4$\cdot10^{-5}$ & 1.6$\cdot10^{-3}$ & 2.6$\cdot10^{-5}$ \cr
                &  val std $u$ & 8.1$\cdot10^{-3}$ & 1.9$\cdot10^{-5}$ & 7.6$\cdot10^{-5}$ & 1.5$\cdot10^{-3}$ & 8.2$\cdot10^{-5}$ \cr
    \cmidrule{1-7}
    Inverse   &  val $\mu$          & 2.7$\cdot10^{-3}$ & 1.5$\cdot10^{-1}$ & 5.1$\cdot10^{-2}$ & 9.1$\cdot10^{-2}$ & 3.7$\cdot10^{-4}$ \cr
    $k = 1$            &  std $\mu$ & 5.0$\cdot10^{-2}$ & 3.6$\cdot10^{-1}$ & 7.2$\cdot10^{-2}$ & 2.1$\cdot10^{-2}$ & 2.5$\cdot10^{-4}$
  \end{tabular}
  \caption{Comparison of the median $L_2$ training and validation loss on Helmholtz's equation against a baseline of manually chosen scalings. The reported values are the median over four independent runs with identical settings. Additionally, we report the standard deviation over the runs of the best performing model on the validation loss.}
  \label{tab:performance_balances_helmholtz}
\end{table*}

\begin{table}[h]
  \centering
    \begin{tabular}{rcccc}
     & GradNorm & LR ann. & SoftAdapt & ReLoBRaLo \\ 
    \midrule  
    $\Delta T_{co} \left[ s \right]$ & 10.4 & 6.7 & 5.0 & 5.2
  \end{tabular}
    \caption{Median computational overhead $\Delta T_{co}$ (in s) per 1'000 optimisation steps compared to using no balancing scheme (4.8s) on Helmholtz's equation.}
  \label{tab:exec_time_helmholtz}
\end{table}

\begin{figure}
	\centering
	\begin{tabular}{c}
		\hline  
		\cellcolor{hellgrau} \small Convergence on inverse Helmholtz problem\\
		\hline  
	\end{tabular} 
	\includegraphics[width=0.8\linewidth]{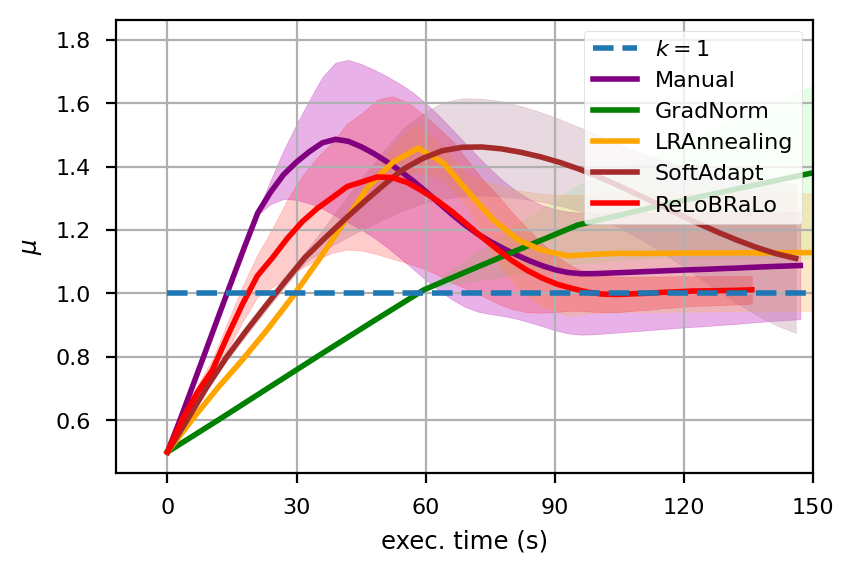}
    \caption{Approximation of the true PDE parameter value $k$ (dashed line) in the inverse problem setting of Helmholtz's equation. Reported values are the mean (solid line) and standard deviation (shaded area) of four independent runs.}
    \label{fig:mu_helmholtz}
\end{figure}
For the inverse Helmholtz problem setting, we select the wave number $k$ to be learned for given data, which we obtained by sampling from the analytically known solution. Furthermore, we initialised $\mu = 0.5$ and tasked the network with approximating $k = 1$. In the Helmholtz inverse problem setting similarly to the inverse Burgers problem, also just one optimiser was chosen for updating the network's parameters $\theta$ together with the PDE parameter $\mu$. ReLoBRaLo also sets a new benchmark for Helmholtz's inverse problem, both in accuracy as well as convergence speed, cf. fig.~\ref{fig:mu_helmholtz} and tab.~\ref{tab:exec_time_helmholtz}.
\begin{table}[h]
  \centering
    \begin{tabular}{lccc}
    Hyperparameter      & Burgers & Kirchhoff & Helmholtz \\ 
    \midrule            
    Learning Rate $l_r$       & $10^{-3}$ & $10^{-3}$ & $10^{-3}$\cr
    Layers $d_K$                    & 4 & 4 & 2\cr
    Neurons per Layer  $w_K$           & 256 & 360 & 256 \cr
    Exponential Decay Rate $\alpha$     & 0.999 & 0.999 & 0.99 \cr
    Temperature $\mathcal{T}$                 & $10^{-1}$ & $10^{-2}$ & $10^{-5}$\cr
    Expected Saudade $\mathbb{E}[\rho]$ & 0.9999 & 0.9999 & 0.99 \cr
    Activation function $\sigma$ & \textit{tanh} & \textit{tanh} & \textit{tanh}
    \end{tabular}
    \caption{Final choices of hyperparameters for architecture and training settings.}
    \label{tab:bayes_opt_parameters_final}
\end{table}
\section{Ablation and Sensitivity Study} \label{sec:Ablation}
The proposed ReLoBRaLo loss balancing scheme together with the PINNs architecture introduce many hyperparameters that have major influence on performance, efficiency and accuracy. In order to investigate the relations between hyperparameters and find their optimal combinations, we conduct an ablation and sensitivity study w.r.t. the temperature $T$, exponential decay rate $\alpha$, and expected saudade $\mathbb{E}[\rho]$ and report its results in this section.
\begin{figure*}
	\centering
	\begin{tabular}{c c c}
		\hline  
		\hspace{6mm}\cellcolor{hellgrau} \small (a) Burgers'  & \hspace{6mm}\cellcolor{hellgrau} \small (b) Kirchhoff's & \hspace{6mm}\cellcolor{hellgrau} \small (c) Helmholtz's \\ 
		\hline  
		\vspace{.5mm}
		\includegraphics[width=0.28\linewidth]{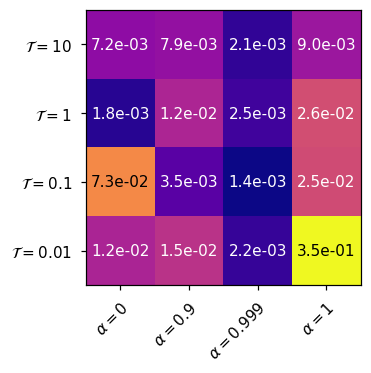} & \includegraphics[width=0.28\linewidth]{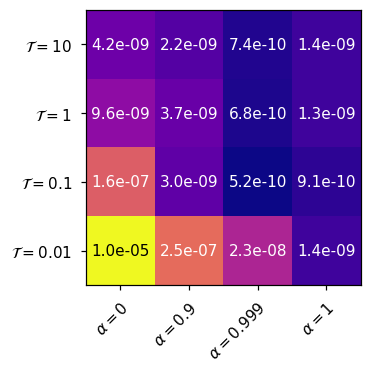}  & \includegraphics[width=0.28\linewidth]{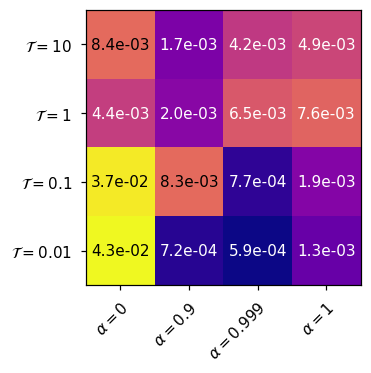} \\
	\end{tabular} 
    \caption{Ablation of the model's performance when varying $\mathcal{T}$ and $\alpha$ with $\mathbb{E}[\rho] = 1$. The reported values are the median of the log $L_2$ loss over multiple training runs.}
    \label{fig:T_alpha_ablation}
\end{figure*}

Fig.~\ref{fig:T_alpha_ablation} visualises the models' sensitivity to the exponential decay rate $\alpha$ and the temperature $T$. A larger $\alpha$ causes the network to "remember" longer, while $T$ controls how much the scalings "sheer out". Fig.~\ref{fig:T_alpha_ablation}(c) shows that Helmholtz's equation benefits most from small values for $T$, which turn the balancing more aggressive. This is in line with the findings in the previous section (cf. sec.~\ref{sec:helmholtz}), where we noted that the large difference in magnitudes between the terms in the loss function caused issues to ReLoBRaLo and that resolute balancing was necessary to avoid the boundary conditions to be neglected. In fact, we found the optimal $T$ to be $10^{-5}$ (cf. tab.~\ref{tab:bayes_opt_parameters_final}). On the other hand, Burgers and Kirchhoff require smoother scalings with a tendency towards higher $T$ and $\alpha$. It is worth noting that all three tasks benefit from the relaxation through the exponential decay, as setting $\alpha = 1$ always causes a deterioration of the model's performance.

However, the relaxation through the exponential decay induces a new trade-off between making the model remember longer and letting it adapt quickly to changes during training. We therefore study the effects of a random lookback through a Bernoulli random variable $\rho$ (\emph{saudade}). It allows setting a lower value for $\alpha$, thus making the model more flexible, while occasionally "reminding" it of its progress since the start of training $\mathcal{L}_i^{(0)}$. 

\begin{table}[h]
  \centering
    \begin{tabular}{lrrr}
    $\mathbb{E}[\rho]$      & Helmholtz & Burgers & Kirchhoff  \\ 
    \midrule            
    0.0   & 2.0$\cdot10^{-3}$ & 1.0$\cdot10^{-3}$ & 1.5$\cdot10^{-09}$ \cr
    0.5   & 5.2$\cdot10^{-5}$ & 9.5$\cdot10^{-3}$ & 2.7$\cdot10^{-09}$ \cr
    0.9   & 4.0$\cdot10^{-5}$ & 1.3$\cdot10^{-3}$ & 2.1$\cdot10^{-09}$ \cr
    0.99  & 2.6$\cdot10^{-5}$ & 4.9$\cdot10^{-4}$ & 6.9$\cdot10^{-10}$ \cr    
    0.999 & 4.1$\cdot10^{-5}$ & 3.8$\cdot10^{-4}$ & 5.6$\cdot10^{-10}$ \cr
    0.9999& 1.2$\cdot10^{-4}$ & 1.4$\cdot10^{-4}$ & 4.0$\cdot10^{-10}$ \cr
    1     & 8.1$\cdot10^{-4}$ & 4.7$\cdot10^{-4}$ & 7.4$\cdot10^{-10}$ \cr
  \end{tabular}
  \caption{Validation loss when varying the expected value of $\rho$. The reported values are the median over three independent runs.}
  \label{tab:performance_lookbacks}
\end{table}

\begin{figure} [h]
	\centering
	\begin{tabular}{c c}
		\hline  
    	\hspace{5mm} \cellcolor{hellgrau} \small (i) $L_2$ Convergence  & \hspace{6mm} \cellcolor{hellgrau} \small (ii) ReLoBRaLo Scaling \hspace{1mm}\\ 
		\hline
	\end{tabular} 
	\includegraphics[width=0.49\linewidth]{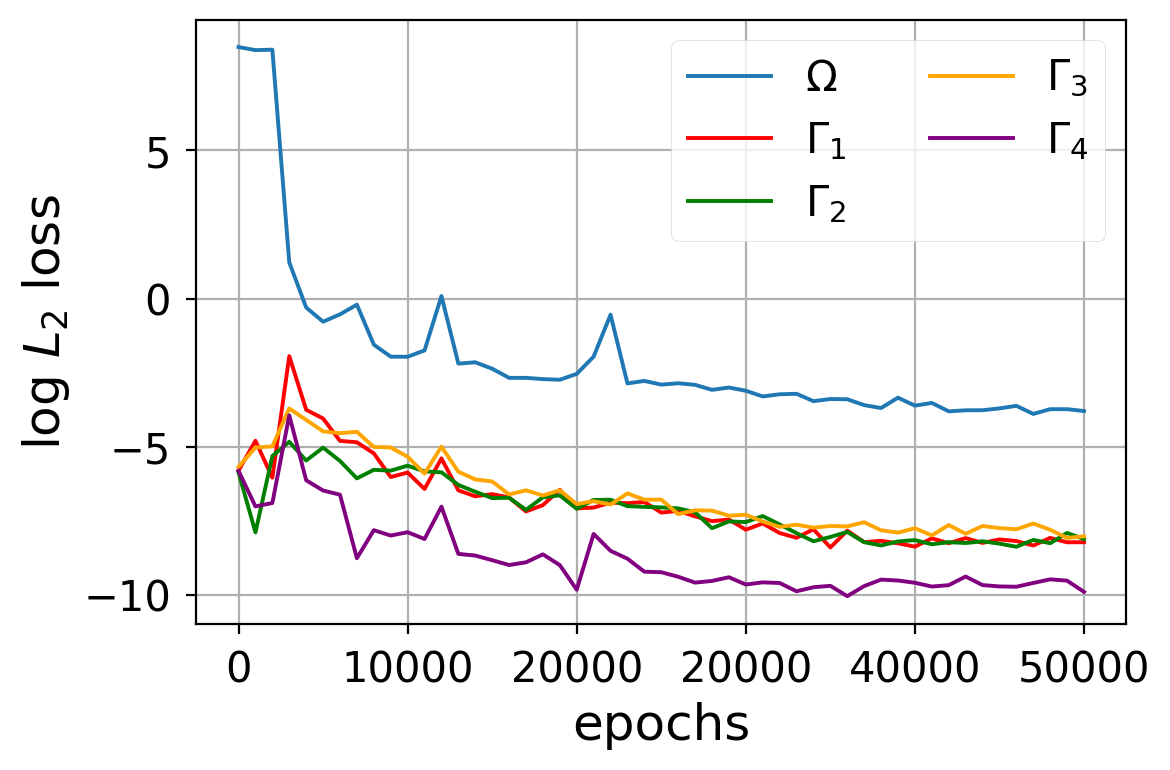} 
	\includegraphics[width=0.4653\linewidth]{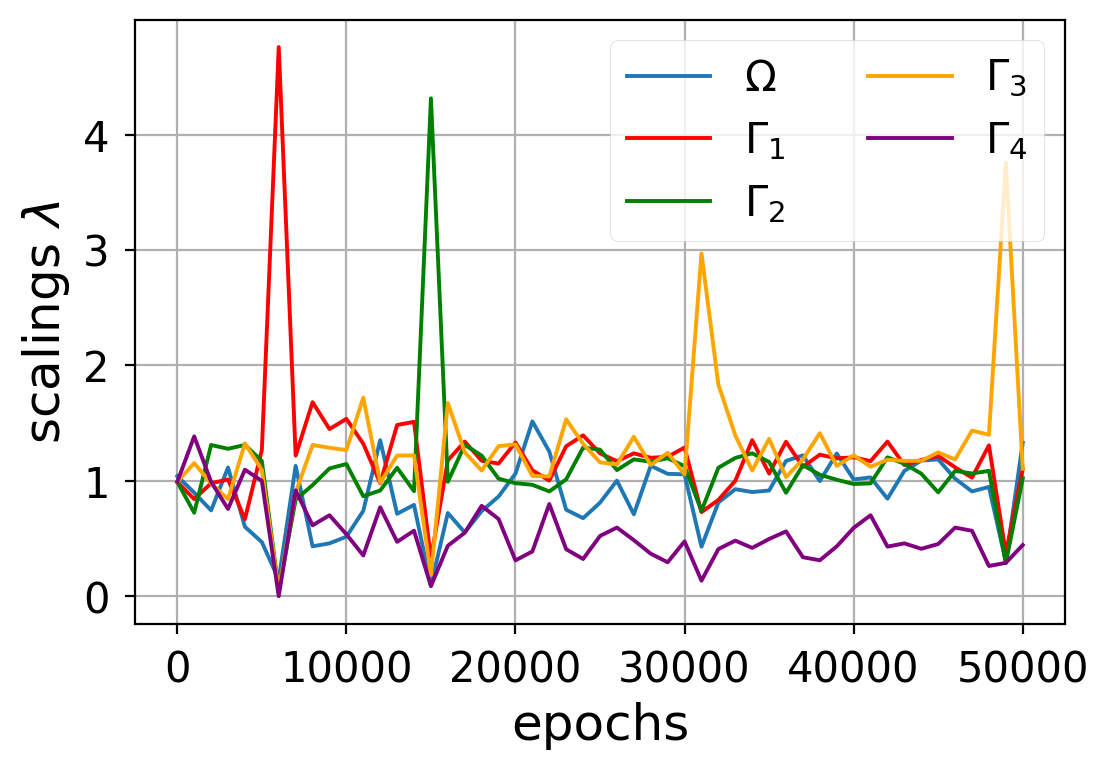}
    \caption{Example of a single training process of ReLoBRaLo on Helmholtz's equation with $\alpha = 0.999$, $\mathcal{T} = 10^{-4}$, $\mathbb{E}[\rho] = 0.9999$.}
    \label{fig:loss_lambda_helmholtz_lookback}
\end{figure}

Tab.~\ref{tab:performance_lookbacks} summarises the change in performance when varying the expected saudade on all three experiments as a comparison. It is apparent that Helmholtz benefits more from frequent lookbacks, as it hits its best performance at  $\mathbb{E}[\rho] = 0.99$, whereas Burgers and Kirchhoff only require an expected lookback every 10'000 optimisation steps. Figs.~\ref{fig:loss_lambda_helmholtz_lookback} and \ref{fig:loss_lambda_burgers} illustrate the effect of random lookbacks. While the stochasticity in the scaling factor increases and therefore makes them less interpretable, it increases the weight on the boundary conditions. The scaled contribution of the boundary conditions consequently leads to a better approximation of the underlying function $\hat{\mathbf{u}}$. 
It is worth noting that the addition of the random lookback improves the accuracy on Helmholtz' equation by more than an order of magnitude while having a lesser, albeit still significant effect on Burgers' and Kirchhoff's equations.

\section{Synopsis and Outlook} \label{sec:summary}
From previous work we observe, that a competitive relationship between physics loss items in the training of PINNs exists and potentially spoils training success, performance or efficiency. This paper investigated different methods aiming at adaptively balancing a loss function consisting of various, potentially conflicting objectives as it may arise in scalarised MOO in PINNs. We proposed a novel adaptive loss balancing method by (i) combining the best attributes of existing approaches, and (ii) introducing a saudade parameter $\rho$ to occasionally incorporate historic loss contribution. This forms a new scheme called \textbf{Relative Loss Balancing with Random Lookback} (ReLoBRaLo) for selecting bespoke weights in order to combine multiple loss terms for the training of PINNs. The effectiveness and merits of using ReLoBRaLo was then demonstrated empirically by investigating several standard PDEs, including solving Helmholtz equation, Burgers' equation and Kirchhoff plate bending equation, and considering both forward problems as well as inverse problems. Our computations showed that ReLoBRaLo is able to consistently outperform the baseline of existing scaling methods (GradNorm, Learning Rate Annealing, SoftAdapt) in terms of accuracy, while also being up to six times more computationally efficient (training epochs or wall-clock time). Finally, we showed that the adaptively chosen scalings $\lambda$ can be inspected to learn about the PINNs training process and identify weak points. This allows to take informed decisions in order to improve the framework.

Future research is concerned with inspection of performance, efficiency, robustness, and scalability of ReLoBRaLo to further PDE classes such as Navier-Stokes equations etc. The adoption of Sobolev Training with Sobolev norms or the incorporation of the Mixture-of-Experts approach \citep{bischof2022mixture} together with ReLoBRaLo may solve the drawback associated with the high costs involved in estimating the neural network solutions of PDEs.


\bibliographystyle{acm}
\bibliography{ReLoBRaLo}

\end{document}